\documentclass[11pt]{article}

\usepackage[final]{acl}

\usepackage{latexsym}
\usepackage{amsthm}
\theoremstyle{plain}
\newtheorem{findings}{Findings}

\usepackage{fontspec}
\usepackage{xunicode}
\usepackage{polyglossia}

\setmainlanguage{english}
\setotherlanguage{hindi}

\newfontfamily\devanagarifont[Script=Devanagari]{FreeSerif.ttf}
\newfontfamily\hindifont[Script=Devanagari]{FreeSerif.ttf}

\usepackage{microtype}

\usepackage{inconsolata}
\usepackage{multirow}

\usepackage{graphicx}
\usepackage{amsmath,amssymb}
\usepackage{mathrsfs} 

\usepackage[table]{xcolor}
\usepackage[most]{tcolorbox}
\usepackage{comment}
\newcommand{\softred}[1]{\textcolor{red!65!black}{#1}}
\newcommand{\softgreen}[1]{\textcolor{green!55!black}{#1}}
\usepackage{hyperref}
\usepackage{array}
\usepackage{tabularx}
\newcolumntype{Y}{>{\centering\arraybackslash}X}
\usepackage{colortbl}
\definecolor{enbg}{gray}{1.00}     
\definecolor{hibg}{gray}{0.96}     
\definecolor{deltabg}{gray}{0.92}   
\definecolor{blockbg}{gray}{1.00} 
\definecolor{lightgray1}{gray}{0.96}

\title{HiMed: Incentivizing Hindi Reasoning in Medical LLMs}

\author{
 \textbf{Dingfeng Jiang\textsuperscript{1,*}},
 \textbf{Han Yan\textsuperscript{1,*}},
 \textbf{Chenze Ma\textsuperscript{1,*}},
 \textbf{Amit Kumar Jaiswal\textsuperscript{2,*}},
 \textbf{Ang Li\textsuperscript{1}},
 \\
 \textbf{Yunxiang Jiang\textsuperscript{1}},
 \textbf{Xinlei Xiong\textsuperscript{1}},
 \textbf{Juhao Liang\textsuperscript{1}},
 \textbf{Hongru Xiao\textsuperscript{1,3}},
 \textbf{Xiang Li\textsuperscript{1,4,$\dagger$}},
 \\
 \textbf{Fan Bu\textsuperscript{1,5}},
 \textbf{Jiale Han\textsuperscript{5}},
 \textbf{Ruchir Gupta\textsuperscript{2}},
 \textbf{Prayag Tiwari\textsuperscript{6,$\dagger$}},
 \textbf{Benyou Wang\textsuperscript{1,4,5,$\dagger$}}
 \\
 \textsuperscript{1}The Chinese University of Hong Kong, Shenzhen,
 \\
 \textsuperscript{2}Indian Institute of Technology (Banaras Hindu University) Varanasi,
 \\
 \textsuperscript{3}Tongji University,
 \textsuperscript{4}Shenzhen Research Institute of Big Data,
 \\
 \textsuperscript{5}Shenzhen Loop Area Institute,
 \textsuperscript{6}Halmstad University
 \\
}

\begin{document}
\maketitle
\renewcommand{\thefootnote}{\fnsymbol{footnote}}
\footnotetext[1]{Equal contribution.}
\footnotetext[2]{Corresponding authors: Xiang Li \href{mailto:lixiang@sribd.cn}{lixiang@sribd.cn}, Prayag Tiwari \href{mailto:prayag.tiwari@ieee.org}{prayag.tiwari@ieee.org}, and Benyou Wang \href{mailto:wangbenyou@cuhk.edu.cn}{wangbenyou@cuhk.edu.cn}}

\setcounter{footnote}{0}
\renewcommand{\thefootnote}{\arabic{footnote}}

\begin{abstract}

Medical large language models hold promise for reducing healthcare disparities, yet Hindi remains severely underrepresented. While medical LLMs excel in high-resource languages, their performance degrades sharply in Hindi, particularly on Indian systems of medicine. We argue that robust cross-lingual medical transfer requires Hindi reasoning. To this end, we introduce \textbf{HiMed}, a Hindi reasoning medical corpus and benchmark suite covering both Western and Indian medicine. We further propose \textbf{HiMed-8B}, a Hindi-form medical reasoning LLM, through the design of \textbf{decaying scaffolding reward}. Extensive experiments demonstrate improvement in Hindi medical reasoning performance and a reduction in the English--Hindi accuracy gap. Ablation studies validate the contribution of each training stage and reward component. All \href{https://github.com/FreedomIntelligence/HiMed}{\texttt{code}}, \href{https://github.com/FreedomIntelligence/HiMed/tree/main/Data}{\texttt{data}}, and \href{https://huggingface.co/collections/FreedomIntelligence/himed}{\texttt{weights}} are available.


\end{abstract}

\section{Introduction}

Large language models (LLMs) have revolutionized medical applications, demonstrating strong performance in high-resource languages \citep{Citrus, patient_education}. Medical decision-making is inherently reasoning-centric: life-critical judgments require causal analysis and factual consistency \citep{DBLP:journals/corr/abs-2411-14461}. Recent studies further show that strengthening reasoning capabilities of medical LLMs can improve decision accuracy \citep{Huatuo-o1,medreason}, underscoring the importance of reasoning.

Beyond accuracy, medical reasoning LLMs also have the potential to reduce healthcare disparities by providing scalable clinical expertise to the underserved across languages \citep{oppo1, oppo2}. However, crossing linguistic boundaries introduces an additional challenge: models must not only transfer reasoning competence, but also align with the linguistic structures and the medical paradigms of the target language \citep{cultureneed1, cultureneed2}.

This challenge is acute when general-purpose models are deployed cross-lingually, as pilot evaluation indicates a hallucination risk. India’s medical ecosystem spans Western and Indian systems of medicine \citep{multineed}. The latter comprise seven officially recognized systems: Ayurveda, Yoga, Naturopathy, Unani, Siddha, Sowa-Rigpa, and Homoeopathy \citep{MoAyush_AyushInIndia_2024}. As many concepts in Indian systems of medicine lack English equivalents \citep{moreno2005dichotomies, nesari2025ayush}, Hindi medical reasoning benefits from operating directly within Hindi representations, making Hindi reasoning necessary.

\begin{figure*}[h]
    \centering
    \includegraphics[width=1\textwidth]{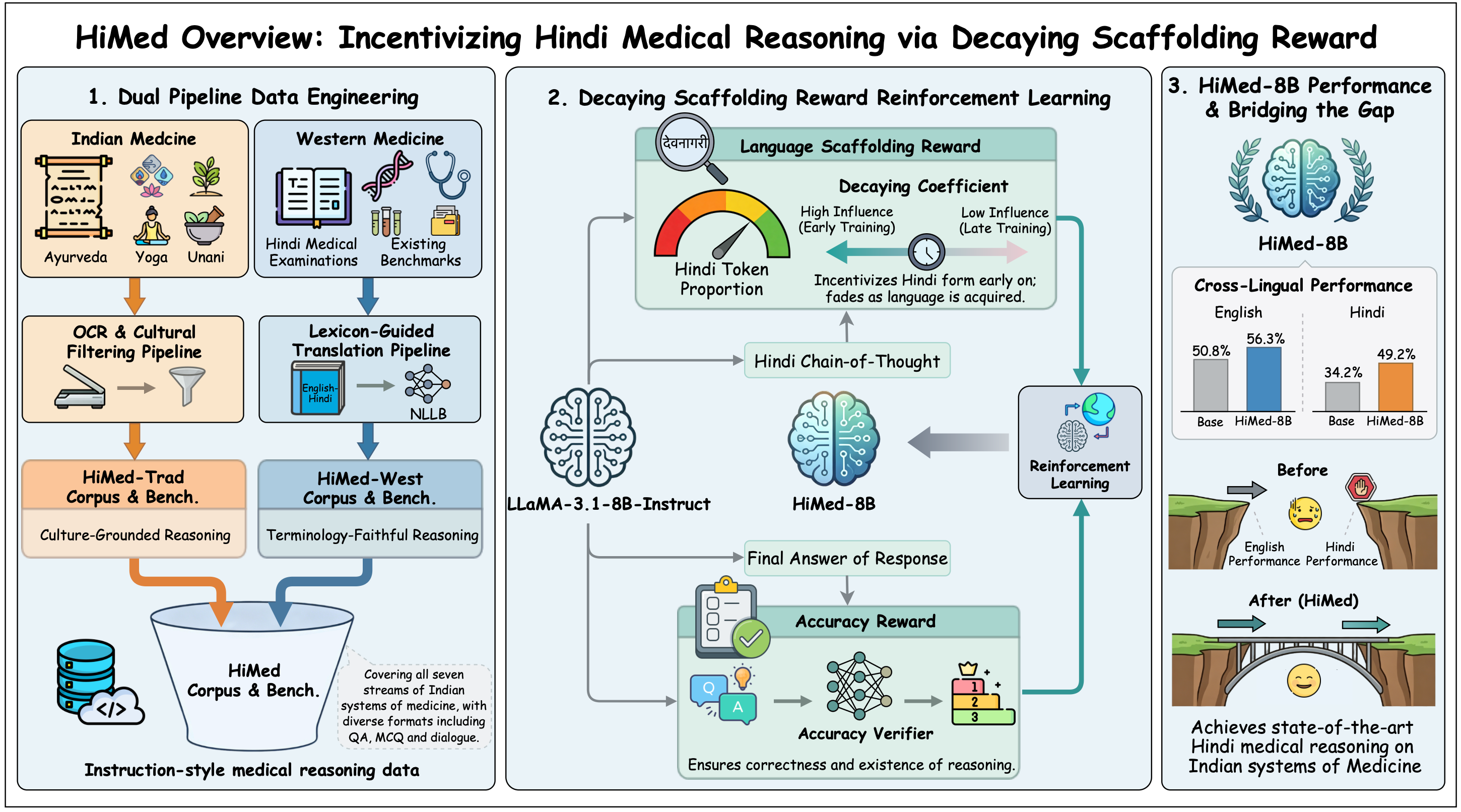}
    \caption{Overview of HiMed and the training framework.}
    \label{fig:main}
\end{figure*}

In this work, we address Hindi medical reasoning at the method and data levels. We propose \textbf{Decaying Scaffolding Reward Reinforcement Learning}, which progressively shifts optimization from reasoning behavior to task-optimal objectives. To support this design, we introduce \textbf{HiMed}, a Hindi medical corpus and benchmark suite covering both Western and Indian systems of medicine. Experiments with LLaMA-3.1-8B-Instruct \citep{llama3} yield the \textbf{HiMed-8B} model, which consistently improves across benchmarks, while ablations validate each training stage and reward component. Figure~\ref{fig:main} provides an overview of the training and evaluation pipeline.

Overall, this work identifies Hindi reasoning as a first-class objective and provides both the data foundation and training methodology. By jointly addressing a series of challenges, our study offers a systematic and extensible framework for adapting general-purpose LLMs to linguistically and culturally specialized medical domains.

\section{Background and Motivations}
\label{sec:bg}
This section motivates Hindi medical reasoning from three progressively stronger angles. We first empirically characterize the English-Hindi medical reasoning gap through a pilot evaluation. We then ask whether this gap can be resolved by translation-based pipelines, and show that translation is insufficient. Finally, we further examine why Hindi medical reasoning remains challenging by identifying two additional bottlenecks at both the data and the algorithm levels.

\subsection{English-Hindi Reasoning Gap }

\noindent \paragraph{Evaluation Setup.} 
Prior work has reported substantial cross-lingual degradation and poor alignment with Indian medical knowledge in LLMs \citep{gap1, gap3, gap4, mixing, ayurbench}. To empirically characterize these gaps, we conduct a pilot evaluation on two representative medical benchmarks reflecting distinct medical paradigms: (i) the medical subsets of MMLU-Pro representing Western medicine, and (ii) RET, an Indian medical examination focusing on Indian systems of medicine. We evaluate GPT-4o \citep{GPT-4o}, HuatuoGPT-o1-8B \citep{Huatuo-o1}, and LLaMA3-8B-UltraMedical \citep{ultramedical} on both benchmarks in English and Hindi. $\Delta$ denotes the accuracy drop from English to Hindi.

\begin{table}[h]
\centering
\footnotesize
\rowcolors{1}{lightgray1}{white}

\newcommand{\modelhead}[1]{{\scriptsize\textsc{#1}}}

\begin{tabular}{lccc}
\hline
\textbf{Benchmark}
& \modelhead{GPT-4o}
& \modelhead{HuatuoGPT-o1}
& \modelhead{UltraMedical} \\
\hline
MMLU-Pro-En        & 60.7\% & 64.7\% & 61.3\% \\
MMLU-Pro-Hi        & 50.7\% & 41.3\% & 34.0\% \\
MMLU-Pro-$\Delta$  & \textbf{10.0\%} & \textbf{23.4\%} & \textbf{27.3\%} \\
\hline
RET-En         & 88.0\% & 38.0\% & 52.0\% \\
RET-Hi         & 70.0\% & 26.0\% & 28.0\% \\
RET-$\Delta$   & \textbf{18.0\%} & \textbf{12.0\%} & \textbf{24.0\%} \\
\hline
\end{tabular}
\caption{Pilot evaluation of cross-lingual performance.}
\label{tab:pilot}
\end{table}

\noindent \paragraph{Results.} As shown in Table~\ref{tab:pilot}, all evaluated models exhibit substantial English–Hindi accuracy drops. On MMLU-Pro, the drops show that even Western medical knowledge does not transfer robustly from English to Hindi. The degradation remains severe for Indian systems of medicine: on RET, open-source models fall to near-random Hindi performance despite relatively stronger English results. These observations suggest that medical reasoning acquired through English-centric training does not consistently generalize to Hindi.

\begin{findings}
\label{find:a}
LLMs exhibit a clear English--Hindi performance gap in Western and Indian medicine.
\end{findings}

\begin{table*}[h] \centering \scriptsize \setlength{\tabcolsep}{1pt} \rowcolors{2}{lightgray1}{white} \begin{tabularx}{\textwidth}{l c c Y Y Y Y Y} \hline \textbf{Resource} & \textbf{Reference} & \textbf{Type} & \textbf{Language} & \textbf{Size} & \textbf{Western Med.} & \textbf{Indian Med.} & \textbf{CoT} \\ \hline MedMCQA & \cite{medmcqa} & Corpus/Benchmark & English & 193K & Yes & No & Yes \\ BioASQ-QA & \cite{bioasq} & Corpus/Benchmark & English & 5K & Yes & No & No \\ PubMedQA & \cite{pubmedqa} & Corpus/Benchmark & English & 273K & Yes & No & No \\ MedNLI & \cite{mednli} & Corpus/Benchmark & English & 14K & Yes & No & No \\ AyurParam & \cite{ayurparam} & Corpus/Benchmark & Hi + En & 4.76M & No & Yes & No \\ ReasonMed & \cite{reasonmed} & Corpus & English & 1.11M & Yes & No & Yes \\ MedReason & \cite{medreason} & Corpus & English & 32K & Yes & No & Yes \\ IndicInstruct & \cite{airavata} & Corpus & Hi + En & 404K & No & No & No \\ UltraMedical & \cite{ultramedical} & Corpus & English & 410K & Yes & No & No \\ AI4Bharat-IndicQA & \cite{ai4bharatIndicQA} & Corpus & Indic & 18K & No & No & No \\ XLingHealth & \cite{claws_lab_xlinghealth_2024} & Corpus & Hi + En & 15K & Yes & No & No \\ MILU-cleaned & \cite{murthyrudra_milu_cleaned_2025} & Corpus & Indic & 74K & Yes & No & No \\ AyurGenixAI & \cite{ayurgenixai} & Corpus & English & 15K & No & Yes & No \\ Multilingual Healthcare & \cite{mlhealthcaredataset} & Corpus & Hi + En & 4K & Yes & No & No \\ MedQA-USMLE & \cite{medqa} & Benchmark & English & 1K & Yes & No & -- \\ MMLU-Pro-Med & \cite{mmlupro} & Benchmark & English & 1K & Yes & No & -- \\ HLE & \cite{hle} & Benchmark & English & 2K & Yes & No & -- \\ HealthBench & \cite{healthbench} & Benchmark & English & 5K & Yes & No & -- \\ GPQA-med & \cite{gpqa} & Benchmark & English & 0.2K & Yes & No & -- \\ MedXpertQA & \cite{medxpertqa} & Benchmark & English & 2K & Yes & No & -- \\ BhashaBench-Ayur & \cite{ayurbench} & Benchmark & Hi + En & 14K & No & Yes & -- \\ \hline \textbf{HiMed-West Corpus} & -- & Corpus & Hindi & 116K & Yes & No & Yes \\ \textbf{HiMed-Trad Corpus} & -- & Corpus & Hindi & 286K & No & Yes & Yes \\ \textbf{HiMed-West Bench} & -- & Benchmark & Hindi & 2K & Yes & No & -- \\ \textbf{HiMed-Trad Bench} & -- & Benchmark & Hindi & 6K & No & Yes & -- \\ \hline \end{tabularx} \caption{Comparison of existing medical corpora and benchmarks.} \label{tab:datasets} \end{table*}

\noindent \paragraph{Implication.} This performance gap could point to a deeper limitation: models may produce fluent Hindi outputs without reasoning in Hindi. Prior work further shows that cross-lingual reasoning degrades without native-language reasoning supervision \citep{xtom, DBLP:journals/corr/abs-2508-14828, DBLP:conf/aaai/ChaiYSGLWLBL0L25}. These observations suggest treating Hindi reasoning as a first-class objective. 

\subsection{Why Hindi Translation Fails}

Translation can fail for two distinct reasons. At the linguistic level, terminology-dense medical content is vulnerable to semantic distortion. At the conceptual level, Indian medical reasoning is grounded in cultural knowledge that is not faithfully recoverable through English reasoning. 

\subsubsection{Reason I: Translation Hallucination}

\noindent \paragraph{Evaluation setup.} Following the observed Hindi reasoning gap, a natural approach is to rely on translation-based pipelines, which translate Hindi inputs into English for reasoning and back into Hindi for delivery. However, such methods rely on translation quality and introduce potential risk, especially in terminology-dense Hindi medical contexts. To quantify this risk, we ask GPT-4o to translate 50 medical sentences into Hindi and ask experts conducting professional post-editing.

\begin{tcolorbox}[
  title=Case I: Translation Hallucination,
  colback=yellow!3, colframe=yellow!50!black, colbacktitle=yellow!8, coltitle=black,
  boxrule=0.6pt,
  left=6pt,
  right=6pt,
  top=6pt,
  bottom=6pt,
  fonttitle=\bfseries
]
\small
\textbf{English Source.}  \\
A 15-year-old \textbf{\softgreen{African-American male}} with a BMI of 22 is brought to his physician by his mother to address concerns about a change in his dietary habits.
  
\vspace{0.5em}
\textbf{GPT Translation.}  \\
\texthindi{एक} 15 \texthindi{वर्षीय \softred{\textbf{आफ्रिकान द्रीपानोसोमा संक्रमणज रोग पुरुष}} जिसका} BMI 22 \texthindi{है, उसकी माता उसे उसके पथ्यसम्बन्धी आदतों में परिवर्तन के बारे में चिंता व्यक्त करने के लिए उसके काया चिकित्सक के पास लाती हैं।} \\
(A 15-year-old \textbf{\softred{African trypanosomal infectious disease male}} with a BMI of 22 is brought by his mother to his body physician in order to express concern about a change in his dietary habits.)
\end{tcolorbox}

\begin{findings}
\label{find:c}
LLM-based translation can introduce semantic hallucinations in terminology-dense Hindi medical content.
\end{findings}

\noindent \paragraph{Results and implication.} We find that 19 out of 50 translated sentences contain translation-induced semantic hallucinations. The above example shows how translation-induced hallucinations can silently corrupt the original semantics. Consequently, Hindi medical LLMs must support Hindi reasoning, allowing valid semantic structures to be formed directly in the target language.

\subsubsection{Reason II: Cultural Grounding}

Hindi reasoning reflects the reality of medical practice in India, where Western and Indian systems of medicine coexist \citep{multineed,chaturvedi2023pluralistic,MoAyush_AyushInIndia_2024}. Medical reasoning is shaped by language. Prior work shows that language-concordant care improves safety and effectiveness \citep{molina2019languageconcordance, reaume2025concordance}. Moreover, many core concepts in Indian systems of medicine lack faithful English equivalents. Existing medical LLMs are predominantly shaped by English-centric training data \citep{pfohl2024equitymedqa}, which limits their ability to reason over Indian systems of medicine in a conceptually faithful manner. Hindi reasoning is not a linguistic preference, but a prerequisite for faithful medical reasoning in the Indian context. 

\begin{findings}
\label{find:d}
Medical reasoning in Indian systems relies on linguistic and cultural grounding that is not preserved under English-centric reasoning.
\end{findings}

\subsection{Why Hindi Reasoning Is Challenging}

Beyond the failure of translation-based approaches, developing effective Hindi medical reasoning models introduces two fundamental bottlenecks: a data bottleneck and an algorithm bottleneck.

\paragraph{Data bottleneck.}
As summarized in Table~\ref{tab:datasets}, existing medical corpora and benchmarks provide limited support for Hindi medical reasoning. Most resources are either English-only or lack coverage of Indian systems of medicine. Consequently, models lack sufficient training signals to learn Hindi medical reasoning behaviors.

\paragraph{Algorithm bottleneck.}
Under accuracy-only reinforcement learning objectives, task-level accuracy provides no supervision on reasoning form: minimal answer-driven rationales and detailed Hindi symptom-to-mechanism reasoning yield identical rewards despite differing substantially.

\begin{tcolorbox}[
  title=Case II: Reasoning Styles,
  colback=yellow!3, colframe=yellow!50!black, colbacktitle=yellow!8, coltitle=black,
  boxrule=0.6pt,
  left=6pt,
  right=6pt,
  top=6pt,
  bottom=6pt,
  fonttitle=\bfseries
]
\small
\textbf{Question.}  \\
\texthindi{एक रोगी में तेज़ बुखार, गर्दन में अकड़न और प्रकाश से डर} \texthindi{पाया गया है। सबसे संभावित निदान क्या है?} \\ 
(A patient presents with high fever, neck stiffness, and photophobia. What is the most likely diagnosis?) \\
(A) \texthindi{मस्तिष्क ट्यूमर} (Brain tumor) \\
(B) \texthindi{मेनिन्जाइटिस} (Meningitis) \\
(C) \texthindi{मिर्गी} (Epilepsy)
(D) \texthindi{माइग्रेन} (Migraine)
\vspace{2pt}

\softred{\textbf{Reasoning A: Direct English conclusion.}}  \\
The correct answer is (B), as these symptoms are consistent with meningitis.
\vspace{2pt}

\softgreen{\textbf{Reasoning B: Hindi symptom-to-mechanism.}}  \\
\texthindi{तेज़ बुखार और गर्दन में अकड़न मस्तिष्कावरण की सूजन का संकेत देते हैं, जबकि प्रकाश से डर इस सूजन के कारण तंत्रिकीय संवेदनशीलता बढ़ने से होता है। इन लक्षणों का संयोजन विकल्प} (B) \texthindi{मेनिन्जाइटिस से मेल खाता है।} \\ (High fever and neck stiffness indicate inflammation of the meninges, while photophobia results from increased neurological sensitivity caused by this inflammation. The combination of these symptoms is consistent with option (B), meningitis.)

\end{tcolorbox}

\begin{findings}
\label{find:e}
Accuracy-only objectives fail to elicit Hindi reasoning form.
\end{findings}

\noindent \paragraph{Implication.} Accuracy-based objectives provide no gradient signal against answer-driven or English rationales, allowing models to converge to objective-equivalent but linguistically misaligned reasoning patterns. Together, Findings~\ref{find:c}, \ref{find:d}, and \ref{find:e} show that Hindi medical reasoning requires both additional data and training objectives that explicitly shape reasoning behavior. This motivates the decaying scaffolding reward and data engineering pipelines introduced in the following section.

\section{Methodology}
\label{sec:method}

Section \ref{sec:bg} suggests that Hindi medical reasoning cannot be reduced to translation, and that accuracy-only learning fails to prefer Hindi reasoning behavior. Our method therefore addresses the problem with objective design and data construction. 

\subsection{Hindi Reasoning Objective}
\label{subsec:hindi_reasoning}

We define Hindi reasoning as a behavioral objective: under Hindi prompts, the model should organize rationales primarily in Hindi, regardless of final answer correctness. We use language form as a lightweight proxy for this objective:
\begin{equation}
\small
R^{\mathrm{lan}}
=
\frac{1}{|y|}
\sum_{t=1}^{|y|}
\mathbb{I}\!\left[y_{t}\ \text{is a Hindi token}\right],
\label{eq:R-lan-proxy}
\end{equation}
where $y$ denotes the generated sequence. $R^{\mathrm{lan}}$ mirrors educational scaffolding \citep{mindinsoc}: supportive structure shapes language form early in training and is gradually withdrawn, allowing the final policy to be governed by task-optimal accuracy. We validate the effectiveness of this proxy through downstream accuracy and human evaluations in Sections~\ref{sec:main} and~\ref{sec:abla}, and Appendix~\ref{app:man_eval}.

\subsection{Decaying Scaffolding Reward Reinforcement Learning (DSR-RL)}
\label{subsec:dsrrl}

\noindent \paragraph{Overview.}
We propose Decaying Scaffolding Reward Reinforcement Learning (DSR-RL), which optimizes the model policy $\pi_\theta(y \mid x)$ using two complementary rewards. For each prompt $x$, we sample a group of $K$ candidate responses $\{y_i\}_{i=1}^K$ from the current policy. To address the lack of supervision over reasoning form under accuracy-only objectives, DSR-RL introduces $R^{\mathrm{lan}}$ to explicitly guide Hindi reasoning behavior along with $R^{\mathrm{acc}}$. A time-dependent coefficient $\lambda(\tau)$ controls the relative influence of the two rewards, enabling early behavioral guidance without constraining the final task-optimal policy. 

\paragraph{Task-optimal reward: reasoning accuracy.}
Each sampled response $y_i$ receives a discrete accuracy reward based on the reasoning behavior and the correctness of the final answer:
\begin{equation}
R^{\mathrm{acc}}_i \in \{0, 0.1, 1\},
\end{equation}
where $0$, $0.1$, and $1$ denote responses without reasoning, responses with reasoning but incorrect answers, and responses with reasoning and correct answers respectively, following prior work~\citep{thinkbeforeanswer, Huatuo-o1}.

\paragraph{Auxiliary scaffolding reward: reasoning language form.}
To encourage Hindi-form reasoning, we define a language-form reward to the entire generated sequence as
\begin{equation}
\small
R^{\mathrm{lan}}_i
=
\frac{1}{|y_i|}
\sum_{t=1}^{|y_i|}
\mathbb{I}\!\left[y_{i,t}\ \text{is a Hindi token}\right]
\;\in\;[0,1].
\label{eq:R-lan}
\end{equation}

\noindent For each reward type $k \in \{\mathrm{acc}, \mathrm{lan}\}$,
we compute a group-relative advantage within $\{y_i\}_{i=1}^K$:
\begin{equation}
\small
\widetilde{R}^{k}_i
=
R^{k}_i
-
\operatorname{mean}_{j\in\{y_i\}_{i=1}^K}\!\left(R^{k}_j\right).
\label{eq:adv-group-mean}
\end{equation}
\begin{equation}
\small
\widehat{R}^{k}_i
=
\frac{\widetilde{R}^{k}_i}
{\operatorname{std}_{j\in\{y_i\}_{i=1}^K}\!\left(\widetilde{R}^{k}_j\right)
+
\varepsilon_{\mathrm{norm}}}.
\label{eq:adv-group-norm}
\end{equation}
The advantage is broadcast to the token level:
\begin{equation}
\small
\widehat{A}^{k}_{i,t} = \widehat{R}^{k}_i,
\qquad
\forall t,\;k \in \{\mathrm{acc}, \mathrm{lan}\}.
\label{eq:adv-broadcast}
\end{equation}
\paragraph{Optimization objectives.}
At each token position $(i,t)$, we compute the likelihood ratio
\begin{equation}
\small
r_{i,t}(\theta)
=
\frac{\pi_\theta(y_{i,t}\mid x,y_{i,<t})}
{\pi_{\theta_{\mathrm{old}}}(y_{i,t}\mid x,y_{i,<t})}.
\label{eq:ratio}
\end{equation}
Using the clipped operator
\begin{equation}
\small
\mathcal{C}(r,A)
=
\min\!\Big(
rA,\;
\operatorname{clip}(r,1-\varepsilon_{\mathrm{clip}},1+\varepsilon_{\mathrm{clip}})\,A
\Big),
\label{eq:clip}
\end{equation}
we define the objective for each reward type as
\begin{equation}
\small
\mathscr{J}^{\mathrm{acc}}
=
\frac{1}{N}
\sum_{i=1}^{N}
\frac{1}{|y_i|}
\sum_{t}
\mathcal{C}\!\left(
r_{i,t}(\theta),
\widehat{A}^{\mathrm{acc}}_{i,t}
\right).
\label{eq:J-acc}
\end{equation}
\begin{equation}
\small
\mathscr{J}^{\mathrm{lan}}
=
\frac{1}{N}
\sum_{i=1}^{N}
\frac{1}{|y_i|}
\sum_{t}
m_{i,t}\,
\mathcal{C}\!\left(
r_{i,t}(\theta),
\widehat{A}^{\mathrm{lan}}_{i,t}
\right),
\label{eq:J-lan}
\end{equation}
where $m_{i,t}\in\{0,1\}$ is a token-level mask that equals $0$ for English medical terms and $1$ otherwise, leveraging principled code-mixing \citep{mixing_is_good}. To prevent excessive policy drift, we include a KL regularization term:
\begin{equation}
\small
\mathrm{KL}
=
\frac{1}{N}
\sum_{i=1}^{N}
\frac{1}{|y_i|}
\sum_{t}
D_{\mathrm{KL}}\!\big(
\pi_\theta(\cdot\mid h_{i,t})
\;\Vert\;
\pi_{\mathrm{ref}}(\cdot\mid h_{i,t})
\big).
\label{eq:KL}
\end{equation}

\paragraph{Final objective.}
The overall reinforcement learning objective is defined as
\begin{equation}
\small
\mathscr{J}(\tau)
=
\bigl(1-\lambda(\tau)\bigr)\,\mathscr{J}^{\mathrm{acc}}
+
\lambda(\tau)\,\mathscr{J}^{\mathrm{lan}}
-
\beta\,\mathrm{KL},
\label{eq:J-final}
\end{equation}
with the corresponding loss
\begin{equation}
\small
\mathscr{L}_{\mathrm{DSR}}(\tau)
=
-\mathscr{J}(\tau).
\end{equation}
Policy optimization is performed using Group Relative Policy Optimization \citep{grpo}. Here, $\lambda(\tau)\in[0,1]$ follows a cosine decreasing function over training iterations, so that auxiliary scaffolding reward dominates early optimization, while task-optimal reward governs the final policy. 

\subsection{Implementations}
\label{subsec:implementations}

We implement the proposed DSR-RL objective via a three-stage training procedure, which progressively prepares the model for reinforcement learning and stabilizes Hindi medical reasoning.

\paragraph{Stage I: language adaptation via SFT (LA).}
The first stage focuses on adapting the model to generate medically grounded Hindi text. This stage stabilizes basic vocabulary and medical expressions, providing a language-aligned initialization for subsequent reasoning-oriented training. Each training example consists of a short factual response without explicit reasoning. 

\paragraph{Stage II: reasoning cold-start via SFT (RC).}
With language generation stabilized, the second stage initializes structured medical reasoning. This stage provides a cold-start that shapes the model’s reasoning behavior. Each training example consists of an explicit chain-of-thought followed by a final answer. For both Stage I and Stage II, we optimize the model using standard supervised fine-tuning with a token-averaged negative log-likelihood. We apply teacher forcing and compute losses only on model output tokens. Sequences longer than the maximum length are truncated, and padding tokens are ignored via attention masks.

\paragraph{Stage III: DSR-RL.}
We perform reinforcement learning according to the design described in Section~\ref{subsec:dsrrl}, with the reference model initialized from the checkpoint obtained at the end of Stage~II.

\subsection{Data Engineering}
\label{sec:data_engineering}

\paragraph{Overview.} We design the HiMed dual data engineering pipelines to construct high-quality reasoning data under two settings: culture-grounded Hindi reasoning for traditional Indian medicine, and terminology-faithful reasoning for Western medicine. Accordingly, HiMed comprises two components: HiMed-Trad and HiMed-West.

\subsubsection{HiMed-Trad}
\paragraph{Objective and feature.} HiMed-Trad is designed to support culture-grounded Hindi reasoning for traditional Indian medicine. It is constructed through a unified OCR-based pipeline tailored for real-world Hindi medical scans. A defining feature of HiMed-Trad is its comprehensive coverage of all seven officially recognized Indian systems of medicine, enabling broad and representative support for culturally grounded medical reasoning.

\paragraph{Data construction pipeline.}
Raw medical documents are first digitized using DeepSeek-OCR~\citep{deepseekocr} and undergo systematic cleaning and segment reconstruction to recover coherent medical passages. We apply LLM-based filtering and calibration, which remove ambiguous content and correct OCR-level artifacts while preserving source fidelity. The resulting high-confidence passages are then converted into MCQ, QA, or dialogue-style instances, with medical rationales that are grounded in the original text.

\paragraph{Evaluation safety and statistics.} HiMed-Trad is divided into two disjoint components: a training corpus and a benchmark. To ensure reliable evaluation, we enforce a strict passage-level split, where all instances derived from the same source passage are treated as an indivisible unit and assigned exclusively to either the corpus or the benchmark, resulting in zero overlap. Dataset statistics are summarized in Table~\ref{tab:himed_trad_compact}, with the full taxonomy and source references provided in Appendix~\ref{app:himed}.

\begin{table}[h]
\centering
\footnotesize
\setlength{\tabcolsep}{12pt} 
\rowcolors{1}{lightgray1}{white}
\begin{tabular}{lcc}
\hline
\textbf{System} & \textbf{Corpus Size} & \textbf{Bench Size} \\
\hline
Ayurveda        & 105,397 & 1000 \\
Yoga            & 8,499  & 1000 \\
Naturopathy     & 20,735  & 1000 \\
Sowa-Rigpa      & 119     & 10 \\
Homoeopathy     & 130,751 & 1000 \\
Unani           & 18,536  & 1000 \\
Siddha          & 2,620   & 1000 \\
\hline
\rowcolor{lightgray1} 
\textbf{Total}  & \textbf{286,657} & \textbf{6,010} \\
\hline
\end{tabular}
\caption{Composition of HiMed-Trad.}
\label{tab:himed_trad_compact}
\end{table}

\subsubsection{HiMed-West}

\paragraph{Objective and feature.}
HiMed-West is designed to support terminology-faithful Hindi reasoning for Western medicine. 
It focuses on accurate handling of standardized medical terminology. To this end, we construct a high-fidelity English--Hindi medical lexicon for the translation pipeline.

\paragraph{Data construction pipeline.}
We adopt a lexicon-guided translation pipeline that combines deterministic terminology injection with machine translation. 
A high-coverage bilingual lexicon is constructed by consolidating authoritative medical glossaries with manual review. 
During translation, English medical terms are first aligned with lexicon entries and preserved, followed by translation using NLLB-200-3.3B~\citep{nllbteam2022languageleftbehindscaling}. 
To facilitate inspection and evaluation, translated medical terms are paired with their original English forms in parentheses. A category-stratified subset of data is translated into Hindi, enabling mixed-language training. We conduct expert audits under a unified rubric to ensure translation fidelity. Compared with NLLB-only and GPT-4o pipelines, the lexicon-guided NLLB approach produces significantly fewer errors and does not exhibit hallucinations in our evaluation. 

\paragraph{Evaluation safety and statistics.}
All translated training corpora are strictly disjoint
from translated benchmarks to prevent leakage.
Data sources and statistics are summarized in
Table~\ref{tab:translation_and_overview}.
GPQA-med refers to the Organic Chemistry, Molecular Biology, and Genetics subsets of GPQA.
HiMed-West Exam is derived from NEET-UG, a national undergraduate medical entrance examination in India.

\begin{table}[h]
\centering
\footnotesize
\setlength{\tabcolsep}{3pt}
\rowcolors{1}{lightgray1}{white}
\begin{tabular}{lrr}
\hline
\textbf{Source} & \textbf{Released Size} & \textbf{Translated Size} \\
\hline
\textbf{HiMed-West Corpus} & \textbf{116,859} & -- \\
\quad MedMCQA        & 182,822 & 91,411 \\
\quad Huatuo-o1 Corpus    & 19,704  &  9,852 \\
\quad MedReason     & 32,682  & 15,596  \\
\textbf{HiMed-West Bench} & \textbf{2,254} & -- \\
\quad MMLU-Pro-Biology      & 717      & 717     \\
\quad MMLU-Pro-Health      &  818    & 818     \\
\quad GPQA-med      &  249 & 249  \\
\quad HiMed-West Exam   &  470 &  -- \\
\hline
\end{tabular}
\caption{Composition of HiMed-West.}
\label{tab:translation_and_overview}
\end{table}

\begin{table*}[t]
\centering
\scriptsize
\setlength{\tabcolsep}{5pt}
\renewcommand{\arraystretch}{1.25}
\begin{tabular}{
l
>{\columncolor{enbg}}c
>{\columncolor{hibg}}c
>{\columncolor{deltabg}}c 
>{\columncolor{enbg}}c
>{\columncolor{hibg}}c
>{\columncolor{deltabg}}c 
>{\columncolor{enbg}}c
>{\columncolor{hibg}}c
>{\columncolor{deltabg}}c 
>{\columncolor{enbg}}c
>{\columncolor{hibg}}c 
>{\columncolor{deltabg}}c 
>{\columncolor{hibg}}c |
>{\columncolor{enbg}}c
>{\columncolor{hibg}}c
>{\columncolor{deltabg}}c
}
\hline
\multirow{2}{*}{\textbf{Model}} &
\multicolumn{3}{c}{\textbf{MMLU-Pro-Bi}} &
\multicolumn{3}{c}{\textbf{MMLU-Pro-He}} &
\multicolumn{3}{c}{\textbf{GPQA-med}} &
\multicolumn{3}{c}{\textbf{HM-W Exam}} &
\multicolumn{1}{c}{\textbf{HM-T}} &
\multicolumn{3}{c}{\textbf{Average}} \\
 & En & Hi & $\Delta$ & En & Hi & $\Delta$ & En & Hi & $\Delta$
 & En & Hi & $\Delta$ & Hi & En & Hi & $\Delta$ \\
\hline

\rowcolor{blockbg}
\multicolumn{17}{l}{\textit{Baseline Models and Our Model}} \\
GPT-4o &  61.9 & \textbf{55.6} & 6.3 & 57.9 & \textbf{47.0} & 10.9 & \textbf{41.8} & 30.5 & 11.3 & \textbf{78.1} & \textbf{78.5} & -0.4 & \underline{62.0} & \underline{59.9} & \textbf{54.7} & 5.2\\ 
UltraMedical-8B & 66.8 & 41.7 & 25.1 & 56.1 & 26.3 & 29.8 & 33.3 & 20.9 & 12.4 & 54.7 & 40.2 & 14.5 & 45.1 & 52.7 & 34.8 & 17.9 \\
MedReason-8B & 65.8 & 43.7 & 22.1 &
58.7 & 27.6 & 31.1 &
27.3 & 29.7 & -2.4 &
53.2 & 41.3 & 11.9 & 47.2 & 51.3 & 37.9 & 13.4\\
HuatuoGPT-o1-8B & 71.0 & 49.1 & 21.9 &
\underline{59.3} & 27.6 & 31.7 &
37.8 & 32.9 & 4.9 &
59.1 & 42.8 & 16.3 &
56.1 & 56.8 & 41.7 & 15.1\\
Qwen2.5-7B-Instruct & \underline{73.1} & 41.8 & 31.3 &
54.4 & 20.3 & 34.1 &
28.9 & 19.7 & 9.2  &
62.0 & 41.5 & 20.5 & 53.3 & 54.6 & 35.3 & 19.3 \\
Qwen2.5-14B-Instruct & 
\textbf{77.7} & 51.9 & 25.8 &
\textbf{62.1}& 26.0 & 36.1 &
33.3& 26.5  & 6.8 &
\underline{71.4}& 46.0 & 25.4 & 
53.2& 
\textbf{61.1} & 40.7 & 20.4 \\
CURE-Med-7B & 
65.8 & 37.9 & 27.9 &
44.0 & 20.8 & 23.2 &
27.7 & 19.7 & 8.0 &
65.8 & 45.8 & 20.0 & 
44.7 & 
50.8 & 33.8 & 17.0 \\
Aya-23-8B & 
49.1& 31.1 & 18.0 &
28.0& 14.8 & 13.2 &
12.4& 10.4 & 2.0 &
38.6& 29.4 & 9.2 & 
46.6& 
32.0& 26.5 & 5.5 \\
AyurParam & 
8.6 & 7.5 & 1.1 &
11.2 & 12.1 & 0.9 &
38.2 & \textbf{43.8} & -5.6 &
27.1 & 28.6 & -1.5 & 
29.8 & 
21.3 & 24.4 & -3.1 \\
LLaMA-3.1-8B-Instruct &
66.7 & 43.5 & 23.2 &
52.6 & 19.7 & 32.9 &
31.3 & 22.5 & 8.8 &
52.6 & 34.3 & 18.3 &
51.0 & 50.8 & 34.2 & 16.6\\
\textbf{HiMed-8B (ours)} & 69.9 & \underline{53.8} & 16.1 &
57.8 & \underline{32.5} & 25.3 &
\underline{38.6} & \underline{36.9} & 1.7 &
58.7 & \underline{46.6} & 12.1 &
\textbf{76.0}  & 56.3 & \underline{49.2} & 7.1\\

\hline
\rowcolor{blockbg}
\multicolumn{17}{l}{\textit{Training Stage Ablation}} \\
w/ LA only & 68.1 & 47.0 & 21.1 & 54.4 & 23.6 & 30.8 &  33.7 & 27.7 & 6.0 & 53.4 & 38.3 & 15.1 & 70.0 & 52.4 & 41.3 & 11.1\\
w/ RC only &
68.3 & 46.2 & 22.1 &
53.1 & 24.7 & 28.4 &
34.9 & 26.9 & 8.0 &
52.8 & 39.6 & 13.2 &
70.3 & 52.3 & 41.5 & 10.8\\
w/ DSR-RL only &
66.7 & 42.0 & 24.7 &
50.5 & 19.8 & 30.7 &
28.9 & 19.3 & 9.6 &
48.4 & 35.0 & 13.4 &
51.3 & 48.6 & 33.5 & 15.1\\
w/o LA &
69.0 & 41.4 & 27.6 &
51.2 & 29.1 & 22.1 &
36.1 & 34.5 & 1.6  &
57.9 & 42.3 & 15.6 &
74.1 & 53.6 & 44.3 & 9.3\\
w/o RC & 69.3 & 46.7 & 22.6 &
51.7 & 19.3 & 32.4 &
36.1 & 33.7 & 2.4  &
51.6 & 34.1 & 17.5 &
69.3 & 52.2 & 40.6 & 11.6\\
w/o DSR-RL & 69.5 & 49.7 & 19.8 & 56.4 & 27.6 & 28.8 & 35.3 & 30.9 & 4.4 & 56.2 & 42.6 & 13.6 & 74.9 & 54.4 & 45.1 & 9.3\\
w/o DSR-RL, full-data SFT & 
69.6 & 50.0 & 19.6 &
56.4& 27.8 & 28.6 &
35.7& 31.7 & 4.0 &
56.0& 42.8 & 13.2 & 
74.9& 
54.4 & 45.4 & 9.0 \\

\hline
\rowcolor{blockbg}
\multicolumn{17}{l}{\textit{Reward Ablation}} \\
w/o $R^{\mathrm{lan}}$ &
70.4 & 51.3 & 19.1 &
58.6 & 30.2 & 28.4 &
38.6 & 32.1 & 6.5 &
59.0 & 44.7 & 14.3 &
75.1 & 56.7 & 46.7 & 10.0\\
w/ binary $R^{\mathrm{acc}}$ & 
69.0 & 52.0 & 17.0 &
56.4 & 30.0 & 26.4 &
37.3 & 35.3 & 2.0 &
57.4& 44.5 & 12.9 & 
74.6& 
55.0 & 47.3 & 7.7 \\
w/ binary $R^{\mathrm{lan}}$ &
66.9 & 47.6 & 19.3 &
53.5 & 24.2 & 29.3 &
34.9 & 30.1 & 4.8 &
54.3 & 38.9 & 15.4 &
72.5 &
52.4 & 42.7 & 9.7 \\
w/ constant reward ratio & 67.5 & 49.0 & 18.5 &
53.8 & 25.4 & 28.4 &
36.1 & 32.5 & 3.6 &
54.0 & 39.6 & 14.4 &
72.0 & 52.9 & 43.7 & 9.2\\
w/ linear decay & 
69.6& 53.1 & 16.5 &
57.2& 31.5 & 25.7 &
38.2& 36.1 & 2.1 &
58.1& 45.5 & 12.6 & 
75.5 & 
55.8 & 48.3 & 7.5 \\

\hline
\end{tabular}
\caption{Unified evaluation across medical reasoning benchmarks. Accuracy (\%) is reported. Within \textit{Baseline Models and Our Model} segment, \textbf{bold} highlights the best scores, and \underline{underlining} indicates the second-best. HM-W and HM-T are short for HiMed-West and HiMed-Trad respectively. ``w/o" and ``w/" denote ``without" and ``with".}
\label{tab:main}
\end{table*}

\subsubsection{Data Quality}

We enforce quality control through multiple expert-audited stages. In the OCR pipeline for HiMed-Trad, two domain experts independently reviewed 200 sampled segments using medical correctness, semantic completeness, Hindi readability, and OCR fidelity as criteria, with 98.0\% judged acceptable. For LLM-driven instruction generation, another expert audit of 200 sampled instances measured source fidelity, medical soundness, and Hindi naturalness, yielding a 99.5\% acceptance rate above the calibrated threshold. In the translation pipeline for HiMed-West, bilingual reviewers audited the lexicon construction and translation pipeline under terminological fidelity, semantic equivalence, format preservation, and linguistic adequacy, and the final pipeline achieved a 97.0\% acceptance rate. Disagreements in all audits are conservatively treated as rejections. All data are constructed from publicly available or properly licensed textbooks, examination papers, and official documents, with no patient records or personal health information involved. We release the resulting resources only under license-compatible settings, and provide details in Appendix \ref{app:datasource}–\ref{app:translation}.

\section{Experiments}

\subsection{Experimental Setup}

\paragraph{Model and training setup.}
HiMed-8B is initialized from LLaMA-3.1-8B-Instruct for comparability with prior medical LLMs with the same backbone (e.g., UltraMedical-8B, MedReason-8B, HuatuoGPT-o1-8B). NLLB is excluded as it targets translation, not reasoning. Training is conducted on a single node equipped with 8 NVIDIA H200 GPUs. Across all stages, the model is trained for approximately 56 hours, including 2 hours for LA, 15 hours for RC, and 39 hours for DSR-RL. Approximately 30\% of the supervision is non-MCQ, providing diverse reasoning signals. Table~\ref{tab:data_composition} summarizes the training data composition. Details of training are in Appendix~\ref{app:details}. 

\begin{table}[h]
\centering
\small
\setlength{\tabcolsep}{8pt} 
\rowcolors{1}{lightgray1}{white}
\begin{tabular}{lccc}
\hline
\textbf{Dataset} & \textbf{LA} & \textbf{RC} & \textbf{DSR-RL} \\
\hline
HiMed-West      & 46.5K & 52K & 10K \\
HiMed-Trad      & 102K & 102K & 10K \\
GSM8K           & --   & 8K & --  \\
DailyDialog     & 11K  & --   & --  \\
Persona-Chat    & 17K  & --   & --  \\
\hline
\end{tabular}
\caption{Composition of training data.}
\label{tab:data_composition}
\end{table}

\paragraph{Evaluation benchmarks.}
We evaluate models on multiple medical benchmarks under the zero-shot setting. Western medicine is assessed with MMLU-Pro-Biology, MMLU-Pro-Health, GPQA-med, and HiMed-West Exam in both English and Hindi. These benchmarks cover complementary aspects of biomedical and clinical reasoning. Indian systems of medicine are evaluated with HiMed-Trad Bench in Hindi, which targets reasoning over concepts and practices specific to the seven Indian systems of medicine. 

\paragraph{Evaluation protocol and metrics.}
Evaluations use a multiple-choice format. In the LA stage, greedy decoding is applied with a multi-class classification head. For all baselines and the RC and DSR-RL stages, responses are decoded through a robust answer-extraction procedure with progressively relaxed matching, ranging from strict pattern detection to similarity-based selection. This procedure is designed to handle variation in response format, ensuring fair evaluation. We apply the same prompts and extraction pipeline to all models within each benchmark. We report accuracy and accuracy drop $\Delta = \mathrm{En} - \mathrm{Hi}$. Results are averaged over three runs to mitigate randomness.

\subsection{Main Results}
\label{sec:main}

\noindent \paragraph{Overall performance.}

Table~\ref{tab:main} reports results across Western and Indian medical reasoning benchmarks. HiMed-8B consistently outperforms open-source baselines in Hindi and substantially narrows the English--Hindi accuracy gap relative to LLaMA-3.1-8B-Instruct, indicating improved cross-lingual transfer. These gains do not come at the cost of English performance: HiMed-8B also improves over the base model in English, as supported by both accuracy results and the human evaluation in Appendix~\ref{app:english_reasoning}. Overall, the proposed framework improves cross-lingual reasoning while better aligning model behavior with Hindi medical usage and domain knowledge.

\noindent \paragraph{Comparison with GPT-4o.}

GPT-4o still exhibits a non-trivial English--Hindi gap, suggesting that model scale alone is insufficient to close the Hindi medical reasoning gap. On HiMed-Trad Bench, HiMed-8B achieves the best performance, surpassing both open-source baselines and GPT-4o. This result indicates that targeted training on Indian systems of medicine can enable an open 8B model to outperform a larger closed-source model in this culturally grounded subdomain.

\noindent \paragraph{Open-ended evaluation.}

To assess whether improvements extend beyond MCQ accuracy, two Hindi-proficient medical experts evaluate 500 open-ended medical questions on accuray and 1--5 Likert scale across three dimensions. As shown in Table~\ref{tab:open_ended_eval}, HiMed-8B consistently outperforms the base model, with inter-annotator Pearson agreement above 0.91 on all dimensions. Details are provided in Appendix~\ref{app:open_eval}.

\begin{table}[h]
\centering
\small
\setlength{\tabcolsep}{4pt}
\rowcolors{1}{lightgray1}{white}
\begin{tabular}{lccc}
\hline
\textbf{Dimension} & \textbf{Base 8B} & \textbf{HiMed-8B} & \textbf{IAA $r$} \\
\hline
Accuracy              & 0.189 & \textbf{0.304} & 0.953 \\
Medical faithfulness  & 3.020 & \textbf{3.966} & 0.922 \\
Reasoning coherence   & 3.072 & \textbf{3.924} & 0.918 \\
Hindi nativeness      & 2.956 & \textbf{3.842} & 0.977 \\
\hline
\end{tabular}
\caption{Open-ended human evaluation.}
\label{tab:open_ended_eval}
\end{table}

\subsection{Ablation Study}
\label{sec:abla}

\noindent \paragraph{Stage-level contribution.}

We ablate each training stage and reward component in DSR-RL. As shown in Table~\ref{tab:main}, the HiMed corpus accounts for about 73\% of the Hindi-side gain over the base model, while DSR-RL contributes the remaining 27\%. Replacing DSR-RL with SFT on the same RL-stage data yields weaker performance, indicating that the gains arise from the optimization procedure rather than data volume alone.

\noindent \paragraph{Reward component analysis.}

Removing any training stage or replacing $R^{\text{lan}}$ with a sparse binary signal degrades Hindi performance, showing that temporal annealing and the two reward terms are complementary. The results further indicate that language-form scaffolding is more effective when gradually withdrawn than kept fixed or sparse.

\noindent \paragraph{Human evaluation.}

To test whether gains reflect more than surface Hindi token substitution, two medical experts conduct blind pairwise comparisons between HiMed-8B and an otherwise identical RL variant trained without the language-form reward. HiMed-8B is preferred in 67.4\% of non-tie cases. Annotators report that preferred responses more often express symptom-to-mechanism reasoning directly in Hindi, rather than English-style rationales wrapped in Hindi tokens. Details are provided in Appendix~\ref{app:human_eval}. Together, three human evaluations provide complementary evidence that the gains correspond to improved Hindi reasoning rather than surface-level language matching.

\noindent \paragraph{Failure analysis.}

A taxonomy-based analysis on the full HiMed-Trad Bench shows that 88.6\% of HiMed-8B's residual failures are knowledge or reasoning errors, while language-mixing failures occur zero times. This pattern suggests that scaffolding with annealing effectively removes surface-level Hindi-form failures. Details are provided in Appendix~\ref{app:failure_analysis}. These results indicate that the language-form scaffolding reward induces qualitatively different reasoning behavior.

\subsection{Effectiveness of Scaffolding Reward}

\begin{figure}[htbp]
  \centering
  \includegraphics[width=1.0\linewidth]{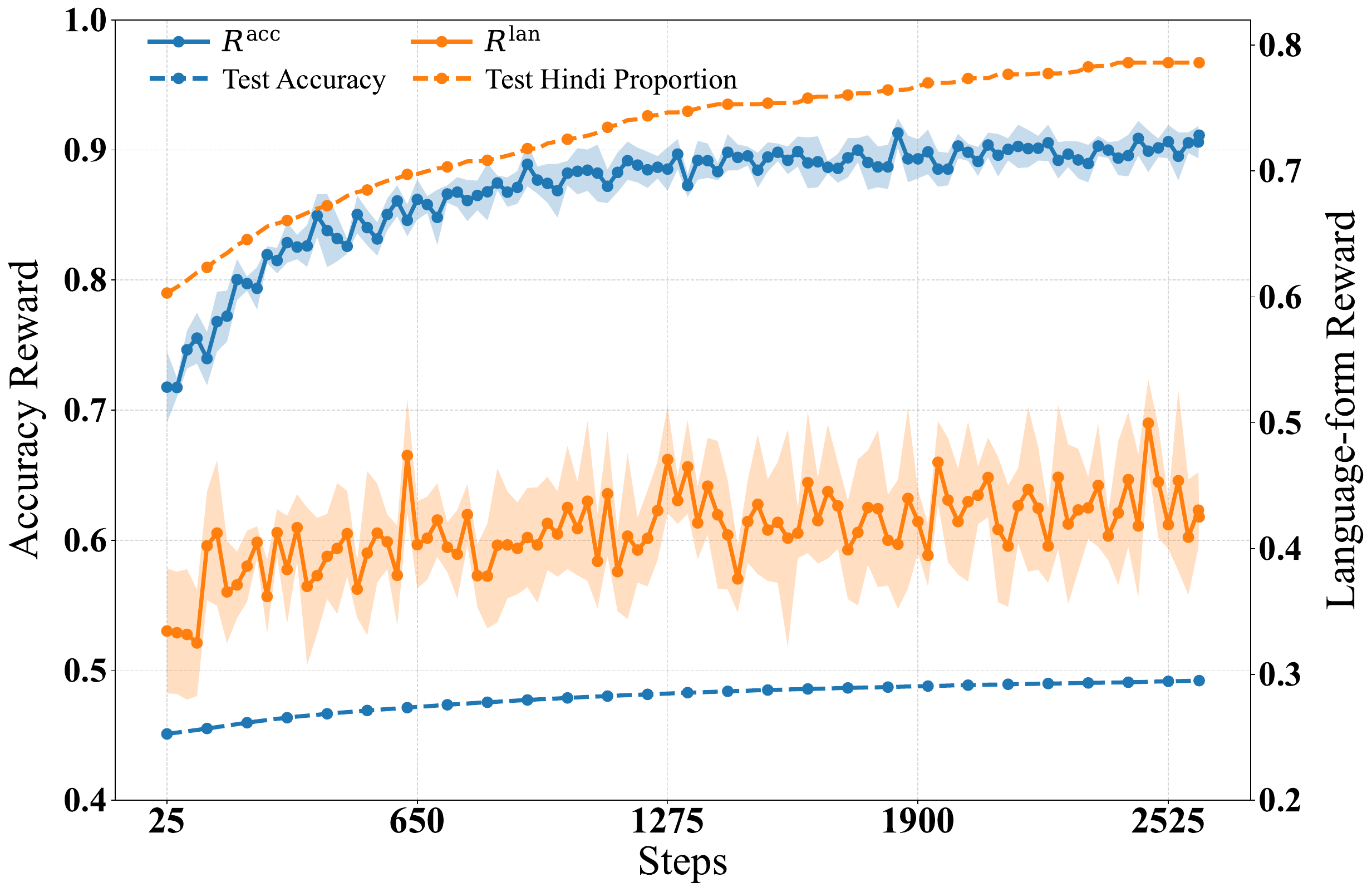}
  \caption{Training dynamics of rewards.}
  \label{fig:reward_dynamics}
\end{figure}

We analyze the training dynamics of the two rewards used in DSR-RL, together with their corresponding test metrics. As shown in Figure \ref{fig:reward_dynamics}, both rewards increase throughout training, and the corresponding test metrics show similar upward trends. In particular, the consistent rise of $R^{\mathrm{lan}}$ and test Hindi proportion indicates that the model progressively produces reasoning chains with higher proportions of Hindi. Meanwhile, the improvement of $R^{\mathrm{acc}}$ and test accuracy suggests that overall reasoning quality is also enhanced. These results confirm that the training-time reward dynamics translate into genuine test-time improvements rather than on-policy reward inflation.

\subsection{Reliability of the Accuracy Reward Model} 

We fine-tune LLaMA-3.2-3B-Instruct on a bilingual medical reasoning dataset as an accuracy verifier. The probability of the True class from the softmax over binary logits is used as the confidence score. A prediction is classified as True if its confidence score exceeds a predefined threshold. On the test set, the verifier achieves 0.969 precision and 0.936 recall. To assess reliability, we use GPT-5 to generate 300 challenging cases by either reversing key logical relations or slightly modifying numerical values. The verifier attains 82.3\% accuracy on this set. Comparing verifier judgments with two medical experts on 300 randomly sampled responses yields 96.7\% exact agreement. Details of the the reward model are in Appendix~\ref{app:reward_training}.

\section{Related Work}

We briefly situate our work within the broader literature and defer a comprehensive discussion to Appendix~\ref{app:relatedwork}. Existing resources for medical language modeling remain predominantly English-centric, with Hindi coverage largely limited to small-scale or translated subsets~\citep{scarcity}. General biomedical corpora and benchmarks, including MedMCQA, PubMedQA, MedNLI, MedQA-USMLE, and MMLU-Pro-Med, have supported progress in medical reasoning but provide limited native Hindi supervision and little coverage of Indian medical systems~\citep{medmcqa, pubmedqa, mednli, medqa, mmlu, mmlupro}. Multilingual resources improve Indic-language coverage, but are typically not specialized for Hindi medical reasoning or rely on translated English-centric data~\citep{airavata, indicllmsuite, gap1, Verma2024}. 

Prior research has investigated cultural alignment~\citep{culturellm, cultureinstruct}, reasoning enhancement~\citep{Creswell2022, debate}, reward modeling~\citep{sbs, rewardsbs}, and large-scale English medical reasoning LLMs~\citep{Huatuo-o1, medpalm2, ultramedical}. However, these directions rarely address the combined requirements of native Hindi reasoning, medical pluralism, and explicit CoT supervision. Closer to our setting, AyurParam~\citep{ayurparam} targets Ayurveda alone with bilingual dialogue supervision but neither releases its training corpus nor includes explicit CoT rationales. BhashaBench-Ayur~\citep{ayurbench} evaluates Ayurveda in Indic languages, but does not cover the broader AYUSH spectrum. Concurrent multilingual reasoning approaches likewise rely on language curricula, translation, distillation, or English-anchored reasoning schemas~\citep{mixing_is_good}, rather than explicitly shaping target-language reasoning behavior.

HiMed differs along three dimensions: it provides an open Hindi corpus with native CoT spanning both Western medicine and all seven Indian systems of medicine, treats Hindi reasoning as a behavioral objective rather than a by-product of translation or instruction transfer, and shapes it through a decaying scaffolding reward design that jointly encourages and balances accuracy, Hindi rationale generation, and format compliance.

\section{Conclusion}

In summary, this work identifies Hindi medical reasoning as a critical yet underexplored requirement for medical LLMs and introduces \textbf{HiMed}, the first large-scale Hindi medical corpus and benchmark suite with native rationales. Building on this foundation, we propose a three-stage adaptation framework centered on \textbf{Decaying Scaffolding Reward Reinforcement Learning}, which yields a concrete instantiation in the form of \textbf{HiMed-8B}. Extensive experiments show that HiMed-8B consistently improves Hindi medical reasoning performance as measured by downstream accuracy, expert preference, and open-ended evaluation across four dimensions, validating the effectiveness of the proposed data and training design. More broadly, our study provides a practical pathway for adapting LLMs to linguistically and culturally specialized medical domains, highlighting a scalable direction toward equitable medical LLMs.

\section*{Limitations}

This work has several limitations that warrant further investigation, particularly with respect to data coverage, model scalability, and real-world applicability, which we outline below. These limitations also highlight important and underexplored directions for future work.

First, while HiMed substantially expands coverage of Hindi medical reasoning, it does not exhaustively represent the full diversity of clinical scenarios or regional variations across India. Certain subdomains, rare conditions, and informal or colloquial clinical expressions may remain underrepresented. In addition, the distribution of training data may introduce biases toward more commonly documented medical language. Future work could improve coverage through more diverse and regionally grounded data sources.

Second, the proposed training framework is instantiated on a single model scale (LLaMA-3.1-8B-Instruct). Although our results demonstrate consistent improvements and narrowed gaps at this scale, it remains unclear how these gains transfer to larger or smaller models. Future work is needed to evaluate scalability, as well as generalization to other low-resource languages beyond Hindi. Extending the framework to larger models and additional languages remains an important direction.

Third, while we emphasize Hindi reasoning and cross-lingual alignment, our evaluation primarily focuses on benchmark-style settings. Real-world clinical interactions may involve more complex, ambiguous, or multi-turn scenarios that are not fully captured in our current evaluation setup. Evaluating in more realistic and interactive clinical settings is an important next step.

Finally, despite the encouraging improvements, our approach does not eliminate the need for clinical expertise, human oversight, or system-level safety mechanisms. HiMed-8B might still produce inaccurate, incomplete, or culturally inappropriate responses, particularly for uncommon or underspecified cases. Performance on educational materials concerning Indian systems of medicine should not be interpreted as evidence of clinical validity or therapeutic effectiveness. As with other medical language models, HiMed-8B is intended for academic research and evaluation purposes only and should not be used for autonomous diagnosis, treatment recommendation, or other safety-critical clinical decision-making.

\section*{Ethical Considerations}

Despite its advanced reasoning capabilities, HiMed-8B may still produce inaccurate outputs. It is therefore not suitable for deployment in real-world clinical settings. We restrict its use to research and evaluation purposes and do not support applications in clinical decision-making or other safety-critical scenarios. Users bear ethical responsibility for respecting these limitations and for avoiding inappropriate use. Performance on benchmarks related to Indian systems of medicine reflects consistency with standardized curricula and examinations. 

We acknowledge that medical knowledge and practice are culturally situated. By including both Western medicine and Indian systems of medicine, our goal is to model how medical knowledge and reasoning are expressed across distinct linguistic, educational, and institutional contexts. Performance on content related to Indian systems of medicine reflects consistency with the source materials and standardized curricula used in this study. The inclusion of traditional medical content does not constitute clinical recommendation.

All datasets are derived from publicly available or licensed sources, including examination materials and textbooks. All our data construction processes comply with ethical standards. No patient records or identifiable individual data are used. Our original code is released under the Apache 2.0 License, and our originally created dataset annotations are released under the CC BY 4.0 License. Third-party source materials remain subject to their original licenses and terms of use; materials that cannot be redistributed are provided through source pointers only. All software packages are provided in the accompanying repository. 

Human annotation and questionnaire procedures involved eight adult bilingual participants from Indian and China, recruited through institutional channels. All participants were proficient in Hindi and English. Before participation, they received written task instructions, were informed that their anonymized responses and judgments would be used for academic research, and provided informed consent. They were compensated at a rate of 400–500 CNY per hour, which exceeded the applicable local professional wage. No direct personal identifiers were retained, and all results are reported only in aggregate. 

\section*{Acknowledgments}

Data collection and travel related to this work was supported by the Jay Chaudhry Software Innovation Centre. This work was supported by Shenzhen Loop Area Institute under Grant 2026SFP004, Shenzhen Medical Research Fund (B2503005),  NSFC grant 72495131,  Major Frontier Exploration Program (Grant No. C10120250085) from the Shenzhen Medical Academy of Research and Translation (SMART), and the International Science and Technology Cooperation Center, Ministry of Science and Technology of China (under grant 2024YFE0203000)，Guangdong Provincial Key Laboratory of Mathematical Foundations for Artificial Intelligence (2023B1212010001)，Program for Guangdong Introducing Innovative and Enterpreneurial Teams, Grant No. 2023ZT10X044，Shenzhen Stability Science Program 2023, Shenzhen Key Lab of Multi-Modal Cognitive Computing，Shenzhen Science and Technology Program ZDSYS20230626091302006. Collectively, this support provided essential resources and a sustained research environment for the development of this work, including data preparation, model training, experimental evaluation, and analysis. 

\bibliography{ref}

\newpage
\appendix

\section*{Appendix}

\section{Pilot Evaluation}
\label{app:pilot}

We randomly sampled 150 test instances from both MMLU-Pro and RET and translated them into Hindi to enable a controlled English–Hindi comparison with identical content, following Appendix~\ref{app:translation}. For MMLU-Pro, only the biology and health subsets were included. RET consists of 100 questions from the 2023 AIAPGET exam, evenly covering four disciplines, and 50 questions from BNYS First Year, Part II examinations. All RET items were standardized as multiple-choice questions prior to translation and evaluation. GPT-4o was prompted to output a single option. HuatuoGPT-o1-8B and LLaMA3-8B-UltraMedical were deployed locally, with predictions determined by the highest-probability option, following their default inference interfaces to avoid confounding prompt engineering. Accuracy is reported as the number of correct predictions out of 150 instances, and $\Delta$ denotes the English-to-Hindi accuracy drop. All evaluations were conducted in a zero-shot setting with single-run results.

\section{Related Work}
\label{app:relatedwork}

\paragraph{Corpora.}
\label{par:datasets}
The dual nature of the Hindi medical ecosystem~\citep{multineed} imposes three requirements for training Hindi medical reasoning LLMs: Hindi coverage, inclusion of Indian medicine, and CoT supervision. No existing corpus satisfies all three. MedMCQA~\citep{medmcqa}, though sourced from NEET PG and AIIMS exams, is predominantly English and focused on Western medicine. Other biomedical resources, including UltraMedical~\citep{ultramedical}, BioASQ-QA~\citep{bioasq}, PubMedQA~\citep{pubmedqa}, MedNLI~\citep{mednli}, ReasonMed~\citep{reasonmed}, and MedReason~\citep{medreason}, are also exclusively English. Multilingual corpora such as IndicInstruct~\citep{airavata}, AI4Bharat-IndicNLP~\citep{ai4bharat}, AI4Bharat-IndicQA~\citep{ai4bharatIndicQA}, IndicLLM-Suite~\citep{indicllmsuite}, XLingHealth~\citep{gap1, claws_lab_xlinghealth_2024}, and MILU-cleaned~\citep{Verma2024, murthyrudra_milu_cleaned_2025} include Hindi but lack medical specialization or rely on translated data. AyurGenixAI~\citep{ayurgenixai} covers Indian medicine yet remains largely English, while the Multilingual Healthcare Text Dataset~\citep{mlhealthcaredataset} provides diverse content without QA or MCQ structure. Furthermore, general-purpose dialogue datasets is insufficient~\citep{persona,dailydialog}. Overall, no corpus jointly supports Hindi, both Western and Indian medicine, and high-quality CoT supervision, leaving a critical gap for downstream training.

\paragraph{Benchmarks.} \label{par:benchmarks}
Benchmarks impose similar requirements of Hindi coverage and inclusion of Indian medicine. Native Hindi evaluation is crucial to avoid translation bias and information loss~\citep{gap1, gap3, apollo}. However, few existing benchmarks meet these criteria. MedQA-USMLE~\citep{medqa}, MMLU-Pro-Med~\citep{mmlu, mmlupro}, HLE-med~\citep{hle}, and HealthBench~\citep{healthbench} are entirely English-based. This limitation extends to advanced biomedical reasoning benchmarks such as GPQA-med~\citep{gpqa} and MedXpertQA~\citep{medxpertqa}. BhashaBench-Ayur~\citep{ayurbench} consists primarily of OCR outputs from examinations, but focuses merely on Ayurveda, leaving a huge gap on other streams. Overall, existing benchmarks remain English-centric and do not support comprehensive evaluation across systems.

\paragraph{Cultural alignment.}
Prior work addresses cultural grounding via region-specific data curation and instruction tuning~\citep{culturellm, cultureinstruct}, or principle-based alignment~\citep{constitution}, alongside culture-aware benchmarks~\citep{cdeval, globe}. However, these efforts primarily target value assessment and remain largely English-centric. Culturally aligned medical reasoning in Hindi, particularly across both Western and Indian medical systems, remains underexplored.

\paragraph{Reasoning enhancement.}
Many techniques have been proposed to improve reasoning, including chain-of-thought prompting~\citep{cot1}, self-consistency~\citep{cot2}, verifier-guided decoding~\citep{Creswell2022}, debate frameworks~\citep{debate}, and process supervision~\citep{sbs}.
These methods improve accuracy on reasoning tasks, but none explicitly addresses the linguistic and cultural fidelity required for Hindi medical reasoning.

\paragraph{Reward modeling.}

Reward modeling for LLM alignment can be grouped into three paradigms: (1) outcome reward models, which score final answers or preferences and underpin RLHF~\citep{humanfeedback}; (2) process reward models (PRMs), which provide rubric or step-level evaluation for reasoning~\citep{sbs, xiao2026structuring} and have been integrated into RL~\citep{rewardsbs}; and (3) LLM-as-a-Judge, which replaces human annotation with strong LLM evaluators for both evaluation and reward generation in RLHF or RLAIF~\citep{LLMaJ1, LLMaJ2, assis, math, bias}. Multi-objective optimization has been explored in mathematical and biomedical reasoning~\citep{grpo, Citrus}, typically via reward combination or structured trade-offs, while distributionally robust optimization~\citep{dro} addresses reward imbalance by optimizing worst-case performance. Despite these rapid advances, medical reward modeling remains largely accuracy-centric, with limited consideration of language nativeness or cultural fidelity.

\paragraph{Medical Reasoning LLMs.}

Recent work has advanced medical reasoning in LLMs through domain-adapted models such as Med-PaLM2, BioGPT, PMC-LLaMA, BioMistral, CURE-Med, AyurParam and HuatuoGPT-o1~\citep{luo2022biogpt, medpalm2, wu2023pmcllama, Huatuo-o1, BioMistral, curemed, ayurparam}, achieving strong performance on benchmarks including USMLE, MedQA, and MMLU~\citep{medqa, mmlu, mmlupro}. These improvements largely stem from three directions: instruction tuning on curated or synthetic medical QA data~\citep{medqa}, reasoning modeling via Chain-of-Thought or structured diagnostic prompting~\citep{cot1}, and preference tuning through preference learning to enhance plausibility and consistency~\citep{Huatuo-o1}. Some general-purpose LLMs also exhibit medical reasoning capabilities \citep{qwen, aya23}.However, existing systems remain predominantly English-centric, with reasoning paradigms grounded in Western clinical practice, and limited exploration of cross-lingual medical reasoning.

\paragraph{Cross-Lingual Transfer.}

Cross-lingual transfer in LLMs is typically achieved through three components: language adaptation, instruction transfer, and representation alignment. Language-adaptive pre-training (LAPT) improves target-language competence via continued pre-training, often combined with vocabulary expansion~\citep{cui2023efficient} or adapter-based embedding learning~\citep{han2025vocadt}. Instruction-following and reasoning are transferred through multilingual instruction data, cross-lingual in-context learning, translation-based methods, and distillation~\citep{wu2025xcit}. Preference alignment has also been extended to multilingual settings by transferring preferences from English-aligned models using implicit rewards or modified DPO on translation-augmented data~\citep{yang2025implicit, lee2025cross}. Finally, representation analyses indicate that middle layers are relatively language-agnostic, motivating explicit middle-layer alignment during fine-tuning~\citep{liu2025middle}.

\paragraph{Cultural Grounding.}
Clinical reasoning in Hindi-speaking regions is shaped by both linguistic variation and India’s institutionalized medical pluralism. While Western clinical practice typically follows a single-paradigm biomedical workflow, healthcare in North and Central India operates within a pluralistic system where biomedicine coexists with traditions (e.g., Ayurveda, Yoga and Naturopathy, Unani, Siddha, Sowa-Rigpa, Homeopathy). This pluralism is structurally embedded in governance, service delivery, and care-seeking, and has been identified as central to India’s pathway toward Universal Health Coverage~\citep{chaturvedi2023pluralistic}. Empirical evidence further shows substantial tradition utilization in outpatient care~\citep{rudra2017ayush}, supported by policy-level institutionalization~\citep{nesari2025ayush}, with qualitative studies highlighting its role in enabling multi-therapy care-seeking~\citep{mallick2016mainstreaming}. This pluralistic ecology entails differences in knowledge organization and reasoning. Indian systems of medicine follow distinct epistemological frameworks not reducible to English biomedical terminology~\citep{nesari2025ayush}, and prior work identifies enduring conceptual divergences from Western biomedicine in framing health, illness, and mind–body relations~\citep{moreno2005dichotomies}. Consequently, faithful clinical reasoning requires operating within the source conceptual system. English-centric reasoning may instead normalize patient narratives into biomedical categories, obscuring clinically relevant choices when patients seek traditional care~\citep{mallick2016mainstreaming, rudra2017ayush}.

\section{Data Sources}
\label{app:datasource}

This section summarizes the data sources used to construct HiMed, including representative books, official documents, and examination papers. The resources span seven Indian systems of medicine: Ayurveda, Homoeopathy, Yoga, Naturopathy, Unani, Siddha, and Sowa-Rigpa. 

\paragraph{Ayurveda.}
Pulse Diagnosis; Aahar Chikitsa; Anubhut Yog; Journal of Ayurveda Case Reports; Essential Drugs List -- Ayurveda; Charak Samhita; Charaka Samhita (Vol. II); Ayurvedic Standard Treatment Guidelines; Rasaraja Mahodadhi; Complete Treatment (Ashtanga Hridayam); Scientific Basis for Ayurvedic Therapies; Sidhparikshapadati Part I; Sushruta Samhita; Ayurvedic Home Remedies; Agada Tantra (Visha Chikitsa); Sushruta Samhita (Sharira Sthana); A Garland of Nectar for Health; Practical Science of Wild Medicinal Herbs.

\paragraph{Homoeopathy.}
Homoeopathic Treatment; Homeopathic Materia Medica (with Repertory); Homeopathic Family Medicine; Illustrated Description of Homeopathic Medicines; Homoeopathy Chikitsa; Lectures on Homoeopathic Philosophy; Saral Homoeopathic Chikitsa; Homeopathy in Animal Diseases; and Saral Homeopathic Ilaj.

\paragraph{Other Systems.}
\textbf{Yoga:} Yoga Therapy; The Yoga Sutras; Indications of Drugs; The Secrets of Yoga. 
\textbf{Naturopathy:} Beneficial for Sexual Health; Why Naturopathy; A New Treatment for Diseases: New Science of Healing; Ankh-Ka: The Art of Naturopathy; Upwash Chikitsa; Health and Hydrotherapy. 
\textbf{Unani:} Fundamental Principles of Unani Medicine; Unani Chikitsasar. 
\textbf{Siddha:} Proven Wild Medicinal Herbs; Proven Remedies (Part I). 
\textbf{Sowa-Rigpa:} Position of Sowa-Rigpa System in Nepalese Buddhist Tradition.

\paragraph{Exams.}
We use publicly available national and university-level examination papers, including NEET-UG (2018, 2020, 2021, 2023, 2024), AIAPGET (Ayurveda, Unani, Siddha), BHU RET (Swasthavritta, Yoga), and BNYS First Year, Part II (University of Patanjali, Dec 2023), covering both Western and Indian systems of medicine. NEET and AIAPGET materials were obtained from the National Testing Agency.\footnote{\url{https://nta.ac.in}} Others were obtained from respective university portals. 

\section{HiMed-Trad Construction}
\label{app:HiMed_Trad}

\subsection{OCR Process}
\label{app:OCR}

We construct HiMed-Trad via a multi-stage OCR pipeline. Merging operations are applied only to non-exam sources, as exam data are already provided at question level. Full prompts are provided in the accompanying repository.

\paragraph{OCR and preprocessing.}
Raw PDFs are converted into high-resolution page images, and text is extracted using DeepSeek-OCR. Outputs are retained as raw MultiMarkdown without manual correction to preserve structural artifacts. A basic cleaning procedure normalizes punctuation and whitespace, while retaining page indices and source identifiers for traceability.

\paragraph{Segment reconstruction.}
To address OCR fragmentation, adjacent text fragments are merged using embedding-based cosine similarity with all-MiniLM-L6-v2~\citep{know1, know_merge1, know_merge2}. Fragments above a threshold are merged, while headings and lists are preserved. Each merged segment is stored with its corresponding page span.

\begin{table*}[h]
\centering
\small
\rowcolors{1}{lightgray1}{white}
\begin{tabular}{lcccccc}
\hline
\textbf{Branch} &
\textbf{Raw} &
\textbf{Cleaned} &
\textbf{Merged} &
\textbf{Filtered} &
\textbf{Calibrated} &
\textbf{Final} \\
\hline
Ayurveda        & 75,164 & 46,040 & 32,420 & 14,424 &  9,085 &  9,054 \\
Yoga            & 11,582 &  5,698 &  3,090 &  1,922 &    844 &    839 \\
Naturopathy     & 17,683 &  7,200 &  3,799 &  2,218 &  1,527 &  1,451 \\
Sowa-Rigpa      &     68 &     57 &     37 &     21 &     10 &     10 \\
Homoeopathy     & 64,752 & 39,953 & 18,400 & 12,673 &  9,316 &  9,226 \\
Unani           &  6,447 &  5,148 &  2,900 &  2,477 &  1,487 &  1,486 \\
Siddha          &  3,125 &  2,122 &  1,385 &    671 &    465 &    453 \\
\hline
\end{tabular}
\caption{Number of Hindi medical segments after each OCR stage.}
\label{tab:trad-branches-stages}
\end{table*}

\paragraph{LLM-based filtering and merging.}
Each segment is classified by GPT-4o along four dimensions: Hindi validity, medical relevance, referential ambiguity, and structural role. Non-Hindi segments are removed. Ambiguous segments are iteratively merged with preceding context until ambiguity is resolved. Only segments satisfying \texttt{Hindi=True, Medical=True, Ambiguity=False, Heading=False} are retained.

\paragraph{Calibration and quality scoring.}
Accepted segments are aligned with page images and corrected for OCR errors using GPT-4o, restricted to character-level fixes without altering meaning. A second LLM pass assigns quality labels (\texttt{NO\_PROBLEM}, \texttt{POSSIBLE\_ISSUE}, \texttt{DEFINITE\_ISSUE}), and only \texttt{NO\_PROBLEM} segments are retained for next stages.

\begin{figure}[h]
    \centering
    \includegraphics[width=1\linewidth]{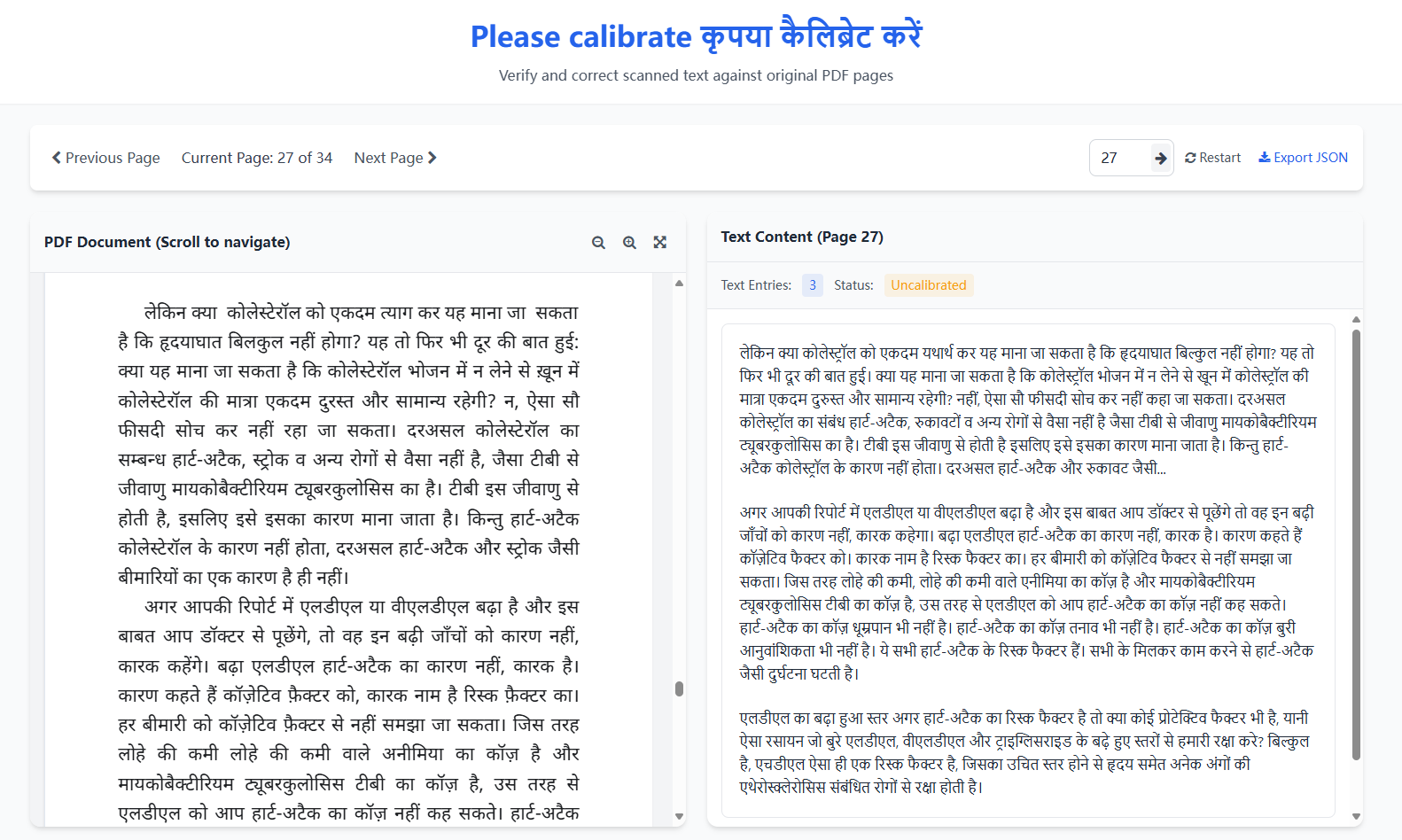}
    \caption{OCR manual inspection UI.}
    \label{fig:placeholder}
\end{figure}

\paragraph{Manual inspection.}
To validate pipeline reliability, two domain experts independently review 200 randomly sampled segments from the retained outputs. Annotators have formal training in medicine or biomedical sciences and are proficient in Hindi medical terminology. Each segment is evaluated along four criteria: (i) \emph{medical correctness}, i.e., factual consistency with standard medical knowledge; (ii) \emph{semantic completeness}, i.e., whether the segment is self-contained; (iii) \emph{coherence and readability}, i.e., whether the text is understandable in Hindi; and (iv) \emph{OCR fidelity}, i.e., whether the content faithfully reflects the source with only minor tolerable errors. Annotators assign a binary decision, and disagreements are conservatively marked as rejection. Across all inspected samples, 98.0\% are judged acceptable, indicating high data quality. Annotators are compensated at 500 CNY per hour, exceeding local translator wage standards, with a total cost of 2{,}000 CNY for this stage. Figure~\ref{fig:placeholder} shows the inspection UI, where scanned pages and OCR outputs are displayed side-by-side.

\subsection{Instruction Pool}
\label{app:instruction}

The instruction pool defines a standardized set of medical question templates used to guide instruction generation. It serves as an intermediate layer between raw medical passages and downstream data generation, bridging the unstructured source content with the structured instruction formats. 

\paragraph{Method.}
The pool is constructed from high-quality medical QA datasets, including HealthBench \citep{healthbench}, HealthSearchQA \citep{Singhal2022}, MedMCQA \citep{medmcqa}, MedReason \citep{medreason}, and ReasonMed \citep{reasonmed}. To integrate heterogeneous sources, GPT-4o abstracts shared intents into standardized instructions. We adopt a five-type taxonomy~\citep{Ely358,Abacha19,DemnerFushman07} to categorize medical questions into \emph{Diagnosis}, \emph{Treatment}, \emph{Etiology}, \emph{Prognosis}, and \emph{Medical Knowledge}, corresponding to core clinical information needs. Instructions are further generalized using placeholder variables derived from SNOMED CT~\citep{SNOMEDCT}, enabling consistent abstraction. To capture realistic user queries, we conduct a bilingual survey on how individuals formulate health-related questions. 

\paragraph{Manual inspection.}
To ensure instruction quality, two domain experts independently review all retained instructions. Annotators have medical or biomedical training and are proficient in Hindi medical terminology. Each instruction is evaluated along two criteria: (i) \emph{clarity}, i.e., whether the intent is well-defined and unambiguous, and (ii) \emph{executability}, i.e., whether the instruction can be reasonably answered using medical reasoning without requiring external context. Annotators assign binary decisions and disagreements are conservatively marked as rejection. Only 61\% of all instructions are retained. Annotators are compensated at 500 CNY per hour, exceeding local professional wage standards, with a total cost of 1{,}000 CNY for this stage. This process ensures that the instruction pool is non-duplicative and suitable for downstream generation. Table~\ref{tab:inscount} summarizes instruction counts by category and source. 
\begin{table}[h]
\centering
\small
\rowcolors{1}{lightgray1}{white}
\begin{tabular}{lccc}
\hline
\textbf{Category} & \textbf{Former Work} & \textbf{Questionnaire} & \textbf{Total} \\
\hline
Diagnosis     & 24 & 7 & 31 \\
Etiology      & 22 & 7 & 29 \\
Prognosis     & 26 & 6 & 32 \\
Treatment     & 26 & 6 & 32 \\
Med.\ Know.   & 26 & 7 & 33 \\
\hline
Total         & 124 & 33 & 157 \\
\hline
\end{tabular}
\caption{Instruction statistics.}
\label{tab:inscount}
\end{table}

\subsection{LLM-Driven Data Generation}
\label{app:datageneration}

We construct HiMed-Trad via a staged pipeline that converts OCR-derived passages into structured instruction data, guided by instruction pool.

\paragraph{Stage I: labeling.}
Each passage is assigned one or more intent categories and instance formats using GPT-4o. Multi-label passages are expanded into atomic entries via the Cartesian product of category and format for simplicity.

\paragraph{Stage II: template-guided generation.}
For each entry, we sample an instruction template from a predefined pool and include few-shot exemplars as style references. Given (\textit{text}, \textit{category}, \textit{format}, \textit{template}), an LLM generates instruction instances. We enforce structured outputs and apply automatic validators to filter malformed generations.

\paragraph{Stage III: quality filtering.}
Generated instances are evaluated by an LLM-based quality rater along four dimensions: contextual grounding, medical correctness, reasoning clarity, and Hindi language quality. Each dimension is scored in a range of $[0,1]$, and instances with an average score below 0.725 are discarded. This threshold is calibrated via a pilot study with domain experts.

\paragraph{Manual inspection.}
We conduct a targeted audit to validate the reliability of automatic filtering. A total of 200 samples are randomly drawn across subjects, formats, and passage types. Two domain experts independently evaluate each instance based on (i) semantic fidelity to the source passage, (ii) medical soundness, and (iii) linguistic naturalness in Hindi. Instances are marked as \emph{pass} only if all criteria are satisfied. Disagreements are conservatively treated as failure. Across all inspected samples, 99.5\% of instances are deemed acceptable, indicating strong agreement between automated filtering and human judgment. Annotators are compensated at 500 CNY per hour, exceeding local professional wage standards, with a total cost of 3{,}000 CNY. Table~\ref{tab:himedtrad_stats} summarizes corpus scale of each format in HiMed-Trad.

\begin{table}[h]
\centering
\small
\rowcolors{1}{lightgray1}{white}
\begin{tabular}{l c}
\hline
\textbf{Format} & \textbf{Count} \\
\hline
QA instances & 93,859 \\
MCQ instances & 102,494 \\
Dialogue instances & 96,314 \\
\hline
\end{tabular}
\caption{Format statistics of HiMed-Trad.}
\label{tab:himedtrad_stats}
\end{table}

\subsection{HiMed-Trad Taxonomy}
\label{app:himed}

\paragraph{Overview.}
In this section, we provide the taxonomy reference of HiMed-Trad Corpus and Bench from seven different streams of Indian systems of Medicine.

\paragraph{Ayurveda.}
We reference the historical overview
\citep{Jaiswal2016Ayurveda} with materials from Directorate
of AYUSH \citep{AYUSH_Delhi_Ashtanga}.

\paragraph{Yoga.}
We rely on evidence from clinical and quality-of-life studies
\citep{Woodyard2011YogaQoL}, and the \textit{Common Yoga Protocol}
periodically issued by the Ministry of Ayush \citep{MoAyush_CYP_2025}.

\paragraph{Naturopathy.}
We include parliamentary records describing its principal modalities,
the Central Council for Research in Yoga and Naturopathy, and WHO's
training benchmarks
\citep{MoAyush_LokSabha_2023Replies,CCRYN_Official,WHO2010NaturopathyBenchmarks}.

\paragraph{Sowa-Rigpa.}
Sowa-Rigpa is represented by the National Institute of Sowa-Rigpa,
the Ministry of Ayush, and pharmacopoeial initiatives by PCIM\&H
\citep{NISR_Official,MoAyush_AyushInIndia_2024,CCRAS_SowaRigpa_2024,PCIMH_Official}.

\paragraph{Homoeopathy.}
We cite the \textit{National Commission for Homoeopathy Act}
\citep{India2020NCHAct} and institutional sources 
\citep{CCRH_Official,NIH_Kolkata_Official}.

\paragraph{Unani.}
Unani medicine is documented through the Ministry of Ayush, CCRUM,
NIUM, and official pharmacopoeial and formulary materials
\citep{MoAyush_AyushInIndia_2024,CCRUM_Official,NIUM_Official}.

\paragraph{Siddha.}
Siddha is covered through the Ministry of Ayush, the National Institute of Siddha
\citep{MoAyush_AyushInIndia_2024,NIS_Official}.

\section{HiMed-West Construction}
\label{app:translation}

We construct HiMed-West via a lexicon-guided translation pipeline designed to ensure terminological fidelity while preserving benchmark structure. Full prompts are provided in the repository.

\subsection{Lexicon Construction}
\label{sec:lexicon}

To guarantee domain-specific precision and natural expression, we build a high-fidelity English--Hindi medical lexicon from both authoritative resources and corpus-driven augmentation.

\paragraph{Authoritative sources.}
The lexicon is initialized from multiple official and educational resources, including \textit{Dictionary of Medicine} \citep{1955dic}, and the NYSED Chemistry Glossary \citep{nysed2018_chemistry_glossary_en_hi}, which together provide standardized terminology and serve as the primary reference.

\paragraph{Digitization and augmentation.}
Scanned volumes are digitized using DeepSeek-OCR with minimal normalization to preserve structural cues. To improve coverage of long-tail terms, we further extract medical entities from the MedReason corpus.

\paragraph{Manual correction and consolidation.}
Two professional bilingual annotators perform OCR correction and translation, focusing on mixed-script and character-level errors, while translations follow deterministic one-to-one mappings with priority given to CSTT standards, yielding a final lexicon of \textbf{45,902} term pairs, with \textbf{62\%} from authoritative sources and \textbf{38\%} from corpus augmentation.

\paragraph{Manual inspection.}
Trained Hindi–English medical translators performed conservative OCR correction and translation without paraphrasing. They evaluated each entry for OCR fidelity and terminological accuracy and excluded ambiguous cases. Annotators were compensated at 400 CNY per hour, above the applicable local professional rate, for a total cost of 30,000 CNY.

\subsection{Translation Pipeline}
\label{sec:translation_method}
We propose a lexicon-guided hybrid pipeline that combines rule-based terminology injection with neural machine translation to ensure both domain fidelity and structural consistency. 

\paragraph{Lexicon-guided translation.}
Medical terms are first matched and replaced with standardized Hindi equivalents from the lexicon, producing a code-mixed intermediate that constrains translation, after which translation is performed using \textbf{NLLB-200-3.3B} \citep{nllbteam2022languageleftbehindscaling}, while retaining original English terms in parentheses.

\paragraph{Format-aware processing.}
To preserve benchmark structure, we adopt an adaptive pipeline: structured MCQs are parsed into components and translated while preserving labels and ordering, whereas unstructured inputs are translated directly. 

\paragraph{Manual comparison inspection.}
We compare three translation pipelines on the same 50 sentences. NLLB+lexicon yields the fewest required fixes and eliminates hallucinations, outperforming both baselines, as shown in Table~\ref{tab:translation_compare_audit}. 

\begin{table}[h]
\centering
\small
\rowcolors{1}{lightgray1}{white}
\begin{tabular}{lccc}
\hline
\textbf{Metric} & \textbf{GPT-4o} & \textbf{NLLB} & \textbf{NLLB+Lexicon} \\
\hline
Need Fix  & 23 & 11 & 2  \\
Hallucination  & 19 & 3 & 0  \\
\hline
\end{tabular}
\caption{Manual audit results.}
\label{tab:translation_compare_audit}
\end{table}

\paragraph{Error analysis.}
We analyze 19 problematic cases and identify four main failure modes. Translation errors are primarily driven by incorrect concept mapping rather than surface fluency, highlighting the importance of lexicon-guided constraints in medical translation settings.

\begin{table}[h]
\centering
\footnotesize
\setlength{\tabcolsep}{6pt}
\rowcolors{1}{lightgray1}{white}
\begin{tabular}{lcc}
\hline
\textbf{Error category} & \textbf{Count} & \textbf{Rate} \\
\hline
Incorrect medical terminology         & 10/19 & 52.6\% \\
Literal but clinically inappropriate  & 9/19  & 47.4\% \\
Semantic hallucination                & 7/19  & 36.8\% \\
Grammatical only                      & 1/19  & 5.3\%  \\
\hline
\end{tabular}
\caption{Error typology.}
\label{tab:err_typology}
\end{table}

\paragraph{Manual final inspection.}
To validate the reliability of the translation pipeline, two bilingual reviewers independently audit sampled instances. Each instance is evaluated along four criteria: (i) terminological fidelity, (ii) semantic equivalence, (iii) format preservation, and (iv) linguistic adequacy. Annotators assign a binary decision. Instances are accepted only if all criteria are satisfied, and disagreements are conservatively rejected. Across an additional audit of 200 samples, the pipeline achieves a 97.0\% acceptance rate. Annotators are compensated at 500 CNY per hour, with total costs of 3{,}000 and 4{,}000 CNY across the two audit stages. 

\section{Data Decontamination and Examples}

\noindent \paragraph{Decontamination.}
To ensure that benchmark performance reflects genuine generalization rather than memorization, we apply multi-level decontamination.
First, \textbf{source-level separation} is enforced by construction: all books and official documents used for corpus generation are strictly disjoint from examination papers used for benchmarking, with no source appearing in both splits.
Second, at the \textbf{passage level}, all data derived from the same OCR passage are assigned exclusively to either the training corpus or the benchmark, preventing partial overlap.
Third, for \textbf{translated benchmarks}, both original questions and their translated or paraphrased variants are explicitly excluded from training data construction, ensuring no indirect exposure during training.

\noindent \paragraph{Examples.}

To provide readers with an intuitive sense of the data format and content in HiMed-Trad Corpus, we present three representative examples covering MCQ, QA, and Dialogue settings. These examples illustrate how traditional medical knowledge is organized into different instruction-following formats while preserving metadata such as subject, system, source, and question identifier.

\newtcolorbox{himedmcqbox}{
  enhanced, breakable,
  colback=green!3, colframe=green!50!black, colbacktitle=green!8, coltitle=black,
  title={HiMed-Trad Corpus Example: MCQ},
  fonttitle=\bfseries, left=2mm, right=2mm, top=1mm, bottom=1mm,
  boxrule=0.5pt, arc=1mm
}

\newtcolorbox{himedqabox}{
  enhanced, breakable,
  colback=blue!3, colframe=blue!50!black, colbacktitle=blue!8, coltitle=black,
  title={HiMed-Trad Corpus Example: QA},
  fonttitle=\bfseries, left=2mm, right=2mm, top=1mm, bottom=1mm,
  boxrule=0.5pt, arc=1mm
}

\newtcolorbox{himeddialoguebox}{
  enhanced, breakable,
  colback=orange!3, colframe=orange!55!black, colbacktitle=orange!10, coltitle=black,
  title={HiMed-Trad Corpus Example: Dialogue},
  fonttitle=\bfseries, left=2mm, right=2mm, top=1mm, bottom=1mm,
  boxrule=0.5pt, arc=1mm
}

\newtcolorbox{himedbox}[1]{
  enhanced, breakable,
  colback=gray!4,
  colframe=gray!55,
  colbacktitle=gray!16,
  coltitle=black,
  title={#1},
  fonttitle=\bfseries,
  left=2mm, right=2mm, top=1mm, bottom=1mm,
  boxrule=0.5pt,
  arc=1mm
}

\begin{himedbox}{HiMed-Trad Corpus Example: MCQ}
\small
\textbf{question\_number:} 00200005012

\textbf{subject:} diagnosis

\textbf{type:} MCQ

\textbf{system:} Ayurveda

\textbf{source:} HiMed-trad

\medskip
\textbf{question:}

\texthindi{एक पारंपरिक चिकित्सक एक रोगी को देख रहे हैं जिसकी नाड़ी लवा की चाल जैसी है। इस संदर्भ में सही निष्कर्ष क्या हो सकता है?}

\medskip
\noindent\textbf{A.} \texthindi{रोगी वात विकार से पीड़ित है}

\noindent\textbf{B.} \texthindi{रोगी के मनोविज्ञान में असंतुलन है}

\noindent\textbf{C.} \texthindi{रोगी सन्निपात से पीड़ित हो सकता है}

\noindent\textbf{D.} \texthindi{रोगी का रत्न परीक्षण गलत था}

\noindent\textbf{E.} \texthindi{रोगी गुरु के दिशा-निर्देशों का पालन नहीं कर रहा}

\textbf{answer:} C

\medskip
\textbf{cot:}

\texthindi{लवा, तीतर, और बटेर की चाल सन्निपात के समय नाड़ी की गति का संकेत होती है। जब नाड़ी लवा की चाल जैसी होती है, तो इसे सन्निपात का संकेत माना जा सकता है। सही निष्कर्ष विकल्प C है।}
\end{himedbox}

\begin{himedbox}{HiMed-Trad Corpus Example: QA}
\small
\textbf{question\_number:} 00200005022

\textbf{subject:} diagnosis

\textbf{type:} QA

\textbf{system:} Ayurveda

\textbf{source:} HiMed-trad

\medskip
\textbf{question:}

\texthindi{एक अनुभवी वैद्य एक मरीज को देखता है जिसकी नाड़ी की गति लवा, तीतर या बटेर की चाल जैसी होती है। मरीज को बुखार है और वह मानसिक भ्रम में है। वैद्य किस प्रकार का प्रकोप या स्थिति निदान करेगा और क्यों?}

\medskip
\textbf{answer:}

\texthindi{सन्निपात, क्योंकि यह नाड़ी की गति और मानसिक भ्रम दोनों के साथ मेल खाता है।}

\medskip
\textbf{cot:}

\texthindi{नाड़ी की गति जो लवा, तीतर या बटेर की चाल जैसी होती है, विशेष रूप से सन्निपात से संबंधित होती है। जब यह गति मानसिक भ्रम के साथ जुड़ी होती है, तो इसे सन्निपात के निदान का संकेत माना जाता है। अनुभवी वैद्य इन लक्षणों के सामूहिक व्याख्या से यह निष्कर्ष निकालता है।}
\end{himedbox}

\begin{himedbox}{HiMed-Trad Corpus Example: Dialogue}
\small
\textbf{question\_number:} 00200005032

\textbf{subject:} diagnosis

\textbf{type:} Dialogue

\textbf{system:} Ayurveda

\textbf{source:} HiMed-trad

\medskip
\textbf{question:}

\textbf{User:} \texthindi{मेरे एक मित्र को कई वर्षों से पेट की समस्या है, और उसे कई बार डॉक्टर ने इलाज के बावजूद सुधार नहीं हुआ। क्या पारंपरिक चिकित्सा में नाड़ी से इस प्रकार की स्थिति का निदान हो सकता है?}

\textbf{Assistant:} \texthindi{पारंपरिक चिकित्सा में नाड़ी का परीक्षण गहन निदान के लिए किया जाता है, इसमें व्यक्ति की नाड़ी की चाल से रोग की प्रकृति को समझा जाता है।}

\textbf{User:} \texthindi{क्या नाड़ी की जाँच से उसके इस लंबे समय से चले आ रहे रोग की जड़ का पता लग सकता है?}

\textbf{Assistant:} \texthindi{हाँ, यदि नाड़ी की चाल का विस्तृत अनुभव है और उसे जौक-सर्प-लवा-तीतर-बटेर की चाल के संदर्भ में सही समझा जाता है।}

\textbf{User:} \texthindi{तो मुझे किस प्रकार के अनुभव की आवश्यकता होगी ताकि मैं नाड़ी से उसके रोग की सही जड़ तक पहुँच सकूँ?}

\medskip
\textbf{answer:}

\texthindi{वर्षों के अभ्यास और गुरु से प्राप्त ज्ञान के माध्यम से नाड़ी की विशिष्ट चालों का प्रत्यक्ष अनुभव आवश्यक है।}

\medskip
\textbf{cot:}

\texthindi{तत्काल ज्ञान के बजाय, दीर्घकालिक अभ्यास और गुरु के मार्गदर्शन से नाड़ी की चालों की सही पहचान और समझ विकसित हो सकती है, जिससे रोग की जड़ की पहचान और निदान सही रूप में संभव हो सके।}
\end{himedbox}

\noindent \paragraph{Structure.}
We organize both the training corpus and benchmarks using a consistent data format.
Each training instance is associated with a stable identifier, a single medical intent category, and one instruction format, and includes a question, its corresponding answer, and a reasoning rationale, all instantiated from predefined templates and grounded in the same source passage.
Benchmark items follow a lighter format, containing only an identifier, the question text, and the gold answer.

\section{Training Details}
\label{app:details}

In this section, we provide details of training, including the verifier model and different training stages of HiMed-8B. For more details, please refer to the GitHub repository. 

\subsection{Reward Model Training Details}
\label{app:reward_training}

We use Llama-3.2-3B-Instruct as the base model. Training data includes 10k Huatuo-o1 English instances and 10k translated Hindi instances, as well as MedReason, 7.5k English and 7.5k Hindi. We generate Chain-of-Thought outputs using LLaMA-3.1-8B-Instruct and obtain supervision labels via GPT-5 by comparing generated and reference reasoning traces for semantic equivalence. Two medical experts independently annotate 300 sampled responses, assessing factual and medical correctness of final answers. Annotations are binary, with emphasis on correctness. Equivalent medical formulations are accepted as well. Annotators are compensated at 500 CNY per hour, higher than the average, totaling 3,000 CNY.

\subsection{SFT Training Details}

We perform supervised fine-tuning for both language adaptation (\textsc{LA}) and reasoning cold-start (\textsc{RC}) using a Llama-3 chat template. Training loss is computed only on assistant tokens, masking prompts with \texttt{-100}. We apply EOS-based dynamic padding without introducing additional pad tokens, and truncate sequences to 4096 tokens for \textsc{LA} and 1536 tokens for \textsc{RC}. Training uses \texttt{Accelerate} and DeepSpeed with bf16 precision and gradient checkpointing, optimized with AdamW and a cosine learning rate schedule with linear warmup. Model selection is based on EMA-smoothed training loss. Table~\ref{tab:LA_RC_hyper} summarizes the hyperparameters used in both LA and RC.

\begin{table}[h]
\centering
\small
\rowcolors{1}{lightgray1}{white}
\begin{tabular}{lcc}
\hline
\textbf{Hyperparameter} & \textbf{LA} & \textbf{RC} \\
\hline
Max sequence length & 4096 & 1536 \\
Micro batch size  & 32 & 8 \\
Gradient accumulation steps & 1 & 2 \\
Global batch size & 256 & 128 \\
Optimizer & AdamW & AdamW \\
Learning rate & 5e-6 & 1e-6 \\
Weight decay & 0.01 & 0.01 \\
Warmup ratio & 0.03 & 0.03 \\
LR scheduler & cosine & cosine \\
Epochs & 3 & 10 \\
Padding strategy & EOS & EOS \\
EMA decay & 0.9 & 0.9 \\
\hline
\end{tabular}
\caption{LA and RC hyperparameter settings.}
\label{tab:LA_RC_hyper}
\end{table}

\subsection{DSR-RL Training Details}

\noindent We adapt the GRPO implementation from the \texttt{trl} framework for reinforcement learning training. Our setup includes (i) a task-optimal reward verifier that has been validated through reliability testing, and (ii) a token-level auxiliary scaffolding reward to provide suitable low-resource language form. For LLaMA-3.1-8B-Instruct, we inject LoRA adapters into the projection modules \texttt{q\_proj}, \texttt{k\_proj}, \texttt{v\_proj}, \texttt{o\_proj}, \texttt{gate\_proj}, \texttt{up\_proj}, and \texttt{down\_proj} in every decoder block for training. The DSR-RL training hyperparameters are summarized in Table~\ref{tab:RL_hyper}, which yielded the best validation performance in our experiments.

\begin{table}[h]
\centering
\small
\rowcolors{1}{lightgray1}{white}
\begin{tabular}{lc}
\hline
\textbf{RL-Hyperparameter} & \textbf{Settings}  \\
\hline
Batch Size      & 8  \\
Learning Rate      &  5e-6  \\
LoRA\_r         & 16   \\
LoRA\_alpha        & 32  \\
LoRA\_dropout           & 0.05    \\
Task-optimal reward threshold           & 0.5     \\
Reward Start Ratio     & 0.1:0.9    \\
Reward End Ratio    & 0.9:0.1    \\
Annealing Strategy    & cosine    \\
GRPO\_clip    & 0.2    \\
KL\_beta    & 0.001    \\
GRPO\_num\_generations & 8 \\
\hline
\end{tabular}
\caption{DSR-RL hyperparameter settings.}
\label{tab:RL_hyper}
\end{table}

\section{Human Evaluation}
\label{app:man_eval}

\subsection{Evaluating English Medical Reasoning}
\label{app:english_reasoning}

\paragraph{Setup.}
We sample 100 English instances and compare HiMed-8B with base model. Two bilingual experts independently assess whether (i) reasoning remains English, (ii) Hindi token insertion or code-mixing occurs, (iii) Hindi-influenced syntax appears, and (iv) reasoning quality degrades. Annotators are compensated at hourly 500 CNY, totaling 2,000 CNY.

\paragraph{Findings.}
HiMed-8B preserves fluent and fully native English reasoning. No Hindi token insertion, Hinglish patterns, or syntactic interference are observed, and no degradation in reasoning quality is identified. These findings align well with quantitative results in Table~\ref{tab:main}.

\paragraph{Case study.}
We provide a representative example showing that HiMed-8B produces fluent English reasoning and improves correctness over the base model, as shown below.

\begin{tcolorbox}[
  title=Case III: English Reasoning Quality,
  colback=yellow!3, colframe=yellow!50!black, colbacktitle=yellow!8, coltitle=black,
  boxrule=0.6pt,
  left=6pt,
  right=6pt,
  top=6pt,
  bottom=6pt,
  fonttitle=\bfseries
]
\small
\textbf{Question.} \\
In reference to a segment of DNA, which of the following molecules contains the fewest number of nucleotides?

\vspace{0.5em}
\textbf{Options.} \\
A: the tRNA transcript from the original DNA \\
B: a single strand of the original DNA segment after a substitution mutation \\
... \\
I: a single strand of complementary DNA (cDNA) made from the original DNA \\
J: a single strand of the original DNA segment after a point mutation

\vspace{0.5em}
\textbf{Ground Truth. I.}

\vspace{0.5em}
\textbf{Base Model (LLaMA-3.1-8B-Instruct).} \\
\textit{``To determine which molecule contains the fewest number of nucleotides, ... the fewest number of nucleotides would be in a single strand of the original DNA segment after a deletion mutation (G).''}

\vspace{0.5em}
\textbf{HiMed-8B.} \\
\textit{``Alright, let’s break this down briefly. ... A processed mRNA has introns removed, so it is shorter than the original gene region. cDNA is made by reverse-transcribing that processed mRNA, ... the best general answer is still the molecule derived from the already-spliced transcript. ... Answer: I. a single strand of complementary DNA (cDNA) made from the original DNA.''}

\end{tcolorbox}

\subsection{Evaluating Open-ended Medical Reasoning}
\label{app:open_eval}

\paragraph{Setup.}
We construct 500 open-ended questions by rewriting MCQs from five benchmarks. For bilingual benchmarks, we include balanced English and Hindi samples. We compare HiMed-8B with LLaMA-3.1-8B-Instruct.

\paragraph{Protocol.}
Two Hindi-proficient medical experts evaluate responses along four dimensions: accuracy, medical faithfulness (1--5 scale), reasoning coherence (1--5 scale), and Hindi nativeness (1--5 scale). Annotators are compensated at 500 CNY per hour, totaling 8,000 CNY. Inter-annotator agreement is high (0.953, 0.922, 0.918, 0.977).

\paragraph{Results.}
HiMed-8B consistently outperforms the base model across all dimensions, achieving higher scores in accuracy (0.304 vs.\ 0.189), medical faithfulness (3.966 vs.\ 3.020), reasoning coherence (3.924 vs.\ 3.072), and Hindi nativeness (3.842 vs.\ 2.956). These results indicate improvements beyond MCQ-style answering.

\subsection{Evaluating Hindi Medical Reasoning}
\label{app:human_eval}

\paragraph{Setup.}
We compare two RL variants initialized from the same Stage-II checkpoint: (i) \textbf{HiMed-8B} (with scaffolding reward) and (ii) \textbf{RL w/o Scaffolding}. All other training settings are identical. We sample 200 prompts covering both Western and Indian medicine. Model outputs are anonymized.

\paragraph{Annotators.}
Two medically trained annotators with native Hindi proficiency independently evaluate responses. Annotators are blinded to model identity and compensated at 500 CNY per hour, totaling 4,000 CNY.

\paragraph{Evaluation protocol.}
Annotators perform blind pairwise comparison using the question: which response exhibits more native Hindi medical reasoning? Judgments focus on reasoning organization rather than fluency, including: (i) Hindi-based reasoning structure, (ii) symptom--mechanism linkage, (iii) avoidance of English-centric reasoning patterns, (iv) coherence, and (v) faithfulness to the prompt. Annotators do not judge factual correctness or penalize principled code-mixing. 

\paragraph{Results.}
Preferences are computed on non-tie cases. HiMed-8B is preferred in 64.5\% and 64.7\% of cases by the two annotators. Using agreement-based aggregation, HiMed-8B is preferred in 67.4\% of non-tie cases, with a tie rate of 32.5\%. Inter-annotator agreement ($\kappa=0.61$) indicates substantial agreement. A two-sided binomial test confirms statistical significance. These results demonstrate that language scaffolding improves native Hindi medical reasoning.

\section{Failure Analysis on HiMed-Trad Bench}
\label{app:failure_analysis}

To better understand the residual gap on Indian systems of medicine, we 
conduct a structured failure analysis on the full HiMed-Trad Bench. 
We first define a six-category failure-mode taxonomy that covers the 
typical error pathways for Hindi medical reasoning, summarized in 
Table~\ref{tab:failure_taxonomy}. We then run HiMed-8B on all 6{,}010 
questions of the HiMed-Trad Bench and ask GPT-5 to classify each of the 
1{,}442 failures into F1--F6, with the option to introduce a new category 
if none fit.

\begin{table}[h]
\centering
\small
\setlength{\tabcolsep}{6pt}
\rowcolors{1}{lightgray1}{white}
\begin{tabular}{cl}
\hline
\textbf{Mode} & \textbf{Description} \\
\hline
F1 & Truncated reasoning  \\
F2 & Wrong terminology mapping \\
F3 & Logical reasoning error \\
F4 & Missing knowledge \\
F5 & Language-mixing failure \\
F6 & Cultural-context misalignment \\
\hline
\end{tabular}
\caption{Failure-mode taxonomy.}
\label{tab:failure_taxonomy}
\end{table}

\paragraph{Failure-mode distribution.} 
Table~\ref{tab:failure_distribution} reports the distribution across all 
1{,}442 failures. F4 dominates at 79.6\%, while F5 is observed exactly zero times.

\begin{table}[h]
\centering
\small
\setlength{\tabcolsep}{6pt}
\rowcolors{1}{lightgray1}{white}
\begin{tabular}{lcc}
\hline
\textbf{Mode} & \textbf{Count} & \textbf{Share} \\
\hline
F4 Missing knowledge          & 1{,}148 & 79.6\% \\
F3 Logical reasoning error    & 130     & 9.0\% \\
F2 Wrong terminology mapping  & 102     & 7.1\% \\
F6 Cultural-context misalignment & 32   & 2.2\% \\
F1 Truncated reasoning        & 25      & 1.7\% \\
Answer-format mismatch$^{*}$  & 5       & 0.3\% \\
F5 Language-mixing failure    & 0       & 0.0\% \\
\hline
\end{tabular}
\caption{Failure-mode distribution on HiMed-Trad Bench. 
$^{*}$Newly introduced during classification.}
\label{tab:failure_distribution}
\end{table}

\paragraph{Per-system failure rate.}
Table~\ref{tab:failure_per_system} reports failure rates by AYUSH system, 
with Yoga and Unani showing the highest rates.

\begin{table}[h]
\centering
\small
\setlength{\tabcolsep}{6pt}
\rowcolors{1}{lightgray1}{white}
\begin{tabular}{lcc}
\hline
\textbf{System} & \textbf{Failures / Total} & \textbf{Rate} \\
\hline
Yoga         & 387 / 1000 & 38.7\% \\
Unani        & 286 / 1000 & 28.6\% \\
Ayurveda     & 223 / 1000 & 22.3\% \\
Siddha       & 199 / 1000 & 19.9\% \\
Naturopathy  & 180 / 1000 & 18.0\% \\
Homoeopathy  & 167 / 1000 & 16.7\% \\
Sowa-Rigpa   & 0 / 10     & 0.0\% \\
\hline
\end{tabular}
\caption{Per-system failure rate.}
\label{tab:failure_per_system}
\end{table}

\paragraph{Findings.}
Three observations follow. First, the residual gap is overwhelmingly a 
knowledge or reasoning issue rather than a language-form issue: F4 and F3 
jointly account for 88.6\% of failures, while F5 is observed zero times. 
This directly supports the claim that the scaffolding-with-annealing 
reward has effectively eliminated surface-level Hindi-form failures, and 
what remains is medical knowledge coverage. Second, failures concentrate 
in Yoga and Unani, where authoritative training corpora 
are sparser and more philosophically heterogeneous than the corresponding 
bench questions. Third, F2 and F6 are small but non-trivial residual 
modes, both natural targets for cross-paradigm reasoning supervision in 
future work.

\section{Prompts and Questionnaire}

\newtcolorbox{promptboxone}{
  enhanced, breakable,
  colback=blue!3, colframe=blue!50!black, colbacktitle=blue!8, coltitle=black,
  title={Prompt: LLM-as-a-Judge Merging},
  fonttitle=\bfseries, left=2mm, right=2mm, top=1mm, bottom=1mm,
  boxrule=0.5pt, arc=1mm
}

\begin{promptboxone}
\small
You are a classification tool. Analyze the following text and output exactly 4 lines:

\medskip
is\_hindi: True/False\\
is\_medical: True/False\\
has\_ambiguity: True/False\\
is\_title\_or\_heading: True/False

\medskip
Definitions:
\begin{itemize}
  \item is\_hindi: True if text is primarily Hindi in Devanagari.
  \item is\_medical: True if it contains medical knowledge (Western/Indian).
  \item has\_ambiguity: True only if the text contains a strong, clearly identifiable, critical referential ambiguity that satisfies all conditions below.
\end{itemize}

Conditions for has\_ambiguity=True:
\begin{enumerate}
  \item The ambiguity concerns a key subject/object/symptom/treatment/causal relation and blocks correct medical interpretation.
  \item The ambiguous element cannot be resolved from the text itself, even with generous natural-language inference.
  \item The ambiguity spans the entire text (the text never provides enough information to clarify the referent).
  \item It is not a normal omission/abbreviation/vagueness/stylistic shortening/common medical phrasing.
  \item It is not a minor unclear phrase that does not affect the main meaning.
\end{enumerate}

If the text is understandable overall (even with minor unclear parts), then has\_ambiguity must be False.

\medskip
Rules:
\begin{itemize}
  \item Output \textbf{only} the 4 lines above.
  \item No explanations.
  \item Use Python-style booleans: True/False.
\end{itemize}

\medskip
Text:
\begin{verbatim}
{text}
\end{verbatim}
\end{promptboxone}

\newtcolorbox{promptboxtwo}{
  enhanced, breakable,
  colback=blue!3, colframe=blue!50!black, colbacktitle=blue!8, coltitle=black,
  title={Prompt: LLM Calibration},
  fonttitle=\bfseries, left=2mm, right=2mm, top=1mm, bottom=1mm,
  boxrule=0.5pt, arc=1mm
}

\begin{promptboxtwo}
\small
You are an OCR post-correction assistant.

You are given OCR text that may span one or more scanned pages. You are also given all corresponding page images.

\medskip
\textbf{Picture:}
\begin{itemize}
    \item The corresponding page images are attached as image files.
\end{itemize}

\medskip
\textbf{Rules:}
\begin{itemize}
    \item Only correct characters, punctuation, spacing, and diacritics.
    \item Do \textbf{not} add new content or delete meaningful content.
    \item Do \textbf{not} translate.
    \item Output \textbf{only} the corrected text.
\end{itemize}

\medskip
OCR text:
\begin{verbatim}
{text}
\end{verbatim}

Page image:
\begin{verbatim}
{image}
\end{verbatim}

\end{promptboxtwo}

\newtcolorbox{promptboxthree}{
  enhanced, breakable,
  colback=blue!3, colframe=blue!50!black, colbacktitle=blue!8, coltitle=black,
  title={Prompt: LLM-as-a-Judge Scoring},
  fonttitle=\bfseries, left=2mm, right=2mm, top=1mm, bottom=1mm,
  boxrule=0.5pt, arc=1mm
}

\begin{promptboxthree}
\small
You are an OCR text quality evaluator.

Your main goal is to determine whether the text contains serious, meaning-breaking OCR corruption.
Default to NO\_PROBLEM unless there is clear evidence of major corruption.

\medskip
Classification rules:

\begin{enumerate}
  \item NO\_PROBLEM:
  \begin{itemize}
    \item The text is readable, understandable, and mostly coherent.
    \item Minor typos, spacing issues, missing matras, imperfect Hindi, or small artifacts are not major OCR problems.
    \item If a normal Hindi reader can infer the meaning easily, choose NO\_PROBLEM.
  \end{itemize}

  \item POSSIBLE\_ISSUE:
  \begin{itemize}
    \item One or two suspicious fragments may be OCR errors, but overall meaning is still understandable.
  \end{itemize}

  \item DEFINITE\_ISSUE:
  \begin{itemize}
    \item The text is seriously corrupted (broken words, incomplete fragments, nonsense sequences) and the meaning cannot be recovered.
  \end{itemize}
\end{enumerate}

\medskip
Output format:
\begin{itemize}
  \item Line 1: exactly one of NO\_PROBLEM / POSSIBLE\_ISSUE / DEFINITE\_ISSUE
  \item Line 2: a short English explanation
\end{itemize}

Do not output Markdown, JSON, or extra text.

\medskip
Text to evaluate:
\begin{verbatim}
{text}
\end{verbatim}
\end{promptboxthree}

\newtcolorbox{promptboxfour}{
  enhanced, breakable,
  colback=blue!3, colframe=blue!50!black, colbacktitle=blue!8, coltitle=black,
  title={Prompt: Forming Instruction},
  fonttitle=\bfseries, left=2mm, right=2mm, top=1mm, bottom=1mm,
  boxrule=0.5pt, arc=1mm
}

\begin{promptboxfour}
\small
You are a professional medical AI data scientist, proficient in multilingual processing. Your task is to analyze medical records and accurately extract the core medical consultation intent. You need to generate standardized outputs in three formats based on the complexity of each case and perform semantic deduplication on instructions sharing the same core intent.

\medskip
\textbf{Processing Steps and Requirements}
\begin{enumerate}
  \item \textbf{Information Analysis and Intent Extraction}
  \begin{itemize}
    \item Carefully read all information points provided by the user, such as symptoms, history, test indicators, doctor's advice, demographics, and location.
    \item Identify and distill the user's core medical consultation intent.
  \end{itemize}

  \item \textbf{Standardized Output Generation}
  \begin{itemize}
    \item Based on question complexity, generate \texttt{Instruction} in one of the following formats:
    \begin{itemize}
      \item \textbf{A. Conversation Summary Format} (multiple user utterances):\\
      Use a narrative summary: ``In the conversation, the user provided [Information 1], [Information 2], ..., ultimately asking [Core Question].''
      \item \textbf{B. QA Standard Format} (one user question):\\
      Use a direct professional question with placeholders (e.g., [symptom], [disease], [procedure]).
      \item \textbf{C. MCQ Format} (requires choice):\\
      Ask in second person and list options (A/B/C/...), using placeholders (e.g., [option]).
    \end{itemize}
  \end{itemize}

  \item \textbf{Intent Classification}
  \begin{itemize}
    \item Assign one primary label: \texttt{Diagnosis}, \texttt{Treatment}, \texttt{Etiology}, \texttt{Prognosis}, or \texttt{Medical Knowledge}.
  \end{itemize}

  \item \textbf{Semantic Deduplication}
  \begin{itemize}
    \item Identify instructions with identical core intent and retain only the clearest, most standard one for each group.
  \end{itemize}

  \item \textbf{Output Format}
  \begin{itemize}
    \item Return a Markdown table with columns: \texttt{Instruction}, \texttt{Category}, \texttt{Form}.
  \end{itemize}
\end{enumerate}

\textbf{Final Output:} Return the Markdown table only.
\end{promptboxfour}

\newtcolorbox{promptbox}{
  enhanced, breakable,
  colback=blue!3, colframe=blue!50!black, colbacktitle=blue!8, coltitle=black,
  title={Prompt: Template-Guided Instruction Generation},
  fonttitle=\bfseries, left=2mm, right=2mm, top=1mm, bottom=1mm,
  boxrule=0.5pt, arc=1mm
}

\begin{promptbox}
\small
\raggedright
\setlength{\parindent}{0pt}
\setlength{\parskip}{2pt}

\textbf{Role}

You are an expert question writer for traditional medicine.

Your task is to generate high-quality instruction-style data
(\texttt{QA}, \texttt{MCQ}, or \texttt{Dialogue})
grounded in traditional medicine texts,
using predefined templates and optional few-shot exemplars.
Ignore all the possible names or personal identifiers.

\medskip
\textbf{Input}

For each generation instance, you are given:
\begin{itemize}
  \item \texttt{text}: source paragraph
  \item \texttt{subject}: medical subject
  \item \texttt{type}: one of \texttt{["MCQ","QA","Dialogue"]}
  \item \texttt{template}: a compatible instruction template
  \item \texttt{few-shot}: optional few-shot exemplars
\end{itemize}

Templates are NOT selected by another model.
Instead, for each \texttt{(subject, type)} pair,
a compatible template is randomly sampled
to reduce selection bias and increase diversity.

\medskip
\textbf{Overall Generation Procedure}

\begin{enumerate}
  \item Select the given template and follow its structure strictly.
  \item Generate three difficulty levels:
  \texttt{EASY}, \texttt{MEDIUM}, and \texttt{HARD}.
  \item Ensure all outputs are self-contained, grounded in the source text,
  and consistent with traditional medicine.
  \item Produce structured outputs that satisfy the required schema.
\end{enumerate}

Malformed generations
(e.g., missing MCQ options, schema violations,
or empty rationales when required)
will be automatically rejected.

\medskip
\textbf{Few-Shot Style References}

The following few-shot blocks are provided as style references only
(do NOT copy sentences verbatim):

\begin{itemize}
  \item \texttt{MCQ\_FEW\_SHOT \{...\}}
  \item \texttt{QA\_FEW\_SHOT \{...\}}
  \item \texttt{DIALOGUE\_FEW\_SHOT \{...\}}
\end{itemize}

\medskip
\textbf{Difficulty Specification}

For each instruction type, you MUST generate all three difficulty levels
in the following fixed order:
\texttt{EASY}, \texttt{MEDIUM}, \texttt{HARD}.

\begin{itemize}
  \item \textbf{EASY}: single fact, no scenario, no reasoning required.
  \item \textbf{MEDIUM}: two-step reasoning, combining about two points;
  brief scenario optional.
  \item \textbf{HARD}: three to five-step reasoning;
  a realistic scenario is REQUIRED
  and must be grounded in traditional medicine.
\end{itemize}

\medskip
\textbf{Type-Specific Instructions}

\textbf{(A) MCQ}

\begin{itemize}
  \item Generate THREE multiple-choice questions
  (\texttt{EASY}, \texttt{MEDIUM}, \texttt{HARD}).
  \item Each question MUST have exactly five options:
  \texttt{A., B., C., D., E.}
  \item There must be ONE correct option.
  \item Provide:
  \begin{itemize}
    \item the correct option letter
    \item a clear reasoning
  \end{itemize}
  \item MCQ is \textbf{NOT} a dialogue:
  \begin{itemize}
    \item Do NOT use \texttt{"User:"} or \texttt{"Assistant:"}
    \item Do NOT create multi-turn conversations
  \end{itemize}
  \item Each item must be a single, self-contained question.
  \item Avoid any mention of \texttt{text}, \texttt{paragraph}, or \texttt{source}.
\end{itemize}

\medskip
\textbf{(B) QA}

\begin{itemize}
  \item Generate THREE short-answer questions
  (\texttt{EASY}, \texttt{MEDIUM}, \texttt{HARD}).
  \item Each item MUST consist of:
  \begin{itemize}
    \item one direct question
    \item one concise correct answer
    \item reasoning
  \end{itemize}
  \item QA is \textbf{NOT} a dialogue:
  \begin{itemize}
    \item NO \texttt{"User:"} or \texttt{"Assistant:"} labels
    \item NO multi-turn format
  \end{itemize}
  \item Avoid any mention of \texttt{text}, \texttt{paragraph}, or \texttt{source}.
\end{itemize}

\medskip
\textbf{(C) Dialogue}

\begin{itemize}
  \item Generate THREE multi-turn dialogues
  (\texttt{EASY}, \texttt{MEDIUM}, \texttt{HARD}).
  \item Each dialogue MUST:
  \begin{itemize}
    \item include two or more turns
    \item contain explicit \texttt{User:} and \texttt{Assistant:} labels
    \item end with a question from the User
  \end{itemize}
  \item Provide:
  \begin{itemize}
    \item the dialogue (up to the final user question)
    \item the correct answer
    \item reasoning
  \end{itemize}
  \item The scenario must clearly belong to traditional medicine.
\end{itemize}

\medskip
\textbf{Output Format}

Return ONLY the following structure
(no explanations, no markdown):

\begin{verbatim}
<EASY><q>...</q><a>...</a><cot>...</cot>
<MEDIUM> ...
<HARD> ...
\end{verbatim}

\begin{itemize}
  \item MCQ: \texttt{<q>} must include options \texttt{A. B. C. D. E.}
  \item QA: must be a single direct question (no dialogue)
  \item Dialogue: must include multi-turn \texttt{User:/Assistant:} format
\end{itemize}

If the task is absolutely impossible, return:
\texttt{<FAIL>}.

\end{promptbox}

\newtcolorbox{raterpromptbox}{
  enhanced, breakable,
  colback=blue!3, colframe=blue!50!black, colbacktitle=blue!8, coltitle=black,
  title={Prompt: Data Quality Rater},
  fonttitle=\bfseries, left=2mm, right=2mm, top=1mm, bottom=1mm,
  boxrule=0.5pt, arc=1mm
}

\begin{raterpromptbox}
\small
\raggedright
\setlength{\parindent}{0pt}
\setlength{\parskip}{2pt}

You are an expert data quality rater for a medical Q-A-CoT dataset in traditional medicine.

You will be given one JSON object with fields:
\begin{itemize}
  \item \texttt{text}: source paragraph (in Hindi)
  \item \texttt{subject}: one of \texttt{["diagnosis","etiology","medical knowledge","prognosis","treatment"]}
  \item \texttt{type}: one of \texttt{["MCQ","QA","Dialogue"]}
  \item \texttt{question}: generated question
  \item \texttt{answer}: generated answer
  \item \texttt{cot}: chain-of-thought reasoning
\end{itemize}

You must evaluate the question/answer/cot on four dimensions, each scored from 0.00 to 1.00:

1. \textbf{grounded\_in\_context}:
\begin{itemize}
  \item Score 1.00 if all information in question/answer/cot is directly derivable from the source text
  \item Score lower if there is any hallucinated information or external knowledge not present in the text
  \item Score 0.00 if the content is completely unrelated to the source text
\end{itemize}

2. \textbf{medical\_correctness}:
\begin{itemize}
  \item Score 1.00 if the medical information is correct according to the source text
  \item Score lower if there are minor inaccuracies or misinterpretations
  \item Score 0.00 if the medical information is incorrect or contradicts the source text
\end{itemize}

3. \textbf{reasoning\_clarity}:
\begin{itemize}
  \item Score 1.00 if the reasoning steps are logical, clear, and well-structured
  \item Score lower if reasoning is somewhat unclear or has minor logical gaps
  \item Score 0.00 if reasoning is illogical, confusing, or missing
\end{itemize}

4. \textbf{language\_quality}:
\begin{itemize}
  \item Score 1.00 if the Hindi language is fluent, natural, and grammatically correct
  \item Score lower if there are minor grammatical errors or awkward phrasing
  \item Score 0.00 if the language is severely broken or incomprehensible
\end{itemize}

Return ONLY a JSON object with 4 float scores in [0.00, 1.00]:

\par\noindent \texttt{\{}
\par\noindent \texttt{\ \ "grounded\_in\_context": 0.00,}
\par\noindent \texttt{\ \ "medical\_correctness": 0.00,}
\par\noindent \texttt{\ \ "reasoning\_clarity": 0.00,}
\par\noindent \texttt{\ \ "language\_quality": 0.00}
\par\noindent \texttt{\}}

Do NOT add any explanations or extra keys. Return only the JSON object.

\end{raterpromptbox}

\newtcolorbox{qbox}{
  enhanced, breakable,
  colback=green!3, colframe=green!50!black, colbacktitle=green!8, coltitle=black,
  title={Questionnaire},
  fonttitle=\bfseries, left=2mm, right=2mm, top=1mm, bottom=1mm,
  boxrule=0.5pt, arc=1mm
}

\begin{qbox}
\small
We are a research team developing AI to better answer health-related questions. Please share a few questions you might ask a doctor or an AI assistant.

\medskip
\texthindi{हम एक शोध टीम हैं जो स्वास्थ्य संबंधी प्रश्नों का बेहतर जवाब देने के लिए} AI \texthindi{(कृत्रिम बुद्धिमत्ता) विकसित कर रहे हैं। कृपया कोई ऐसे प्रश्न साझा करें जो आप एक डॉक्टर या} AI \texthindi{सहायक से पूछ सकते हैं।}

\begin{enumerate}
  \item Imagine you wake up feeling unwell. You have sudden blurred vision in your left eye. You decide to message a doctor or a reliable AI health assistant for advice.

  How would you describe your symptoms to a doctor or AI to get help with diagnosis?

  \medskip
  \texthindi{कल्पना कीजिए कि आप बीमार महसूस करते हुए उठते हैं। आपकी बाईं आंख में अचानक धुंधला दिखाई देने लगा है। आप सलाह के लिए डॉक्टर या एक विश्वसनीय} AI \texthindi{स्वास्थ्य सहायक को मैसेज करने का फैसला करते हैं।}

  \texthindi{निदान में मदद पाने के लिए आप डॉक्टर या} AI \texthindi{को अपने लक्षण कैसे बताएंगे?}

  \item After your recent physical examination at your company/school, you received your report, which showed that several indicators (e.g., white blood cells, blood pressure) were beyond the normal range. You are worried. If you want to ask the doctor/AI why your results are abnormal and what might be causing these changes, what would you say?

  \medskip
  \texthindi{आपकी कंपनी/स्कूल में हाल ही में हुई स्वास्थ्य जाँच के बाद आपको अपनी रिपोर्ट मिली, जिसमें कई जाँच संबंधी मान (जैसे श्वेत रक्त कोशिकाओं की संख्या और रक्तचाप) सामान्य सीमा से बाहर पाए गए। आप चिंतित हैं। यदि आप डॉक्टर या} AI \texthindi{से यह पूछना चाहते हैं कि आपकी जाँच के परिणाम असामान्य क्यों हैं और इन बदलावों के क्या कारण हो सकते हैं, तो आप क्या कहेंगे?}

  \item If you unfortunately contract chikungunya, a mosquito-borne viral disease, which causes persistent fever, and you are very worried that you may not live long or have serious long-term effects, what would you ask the doctor?

  \medskip
  \texthindi{यदि आप दुर्भाग्य से चिकनगुनिया बुखार (एक नया वायरस) की चपेट में आ जाते हैं, जिससे लगातार बुखार आता है, और आप बहुत चिंतित हैं कि कहीं आपकी उम्र कम न हो जाए या गंभीर दीर्घकालिक प्रभाव न हों, तो आप डॉक्टर से क्या पूछेंगे?}

\end{enumerate}
\end{qbox}

\section{Use of AI Assistance}

AI assistants were used in a limited capacity during the preparation of this manuscript. Specifically, large language models were employed to support improvements in clarity and readability of portions of the English text. Research models (GPT-4o, GPT-5, and LLaMA-3.1-8B-Instruct) were used for data construction, labeling, challenge-set generation, and error classification. All substantive aspects of the work, including research conception, methodological design, experimental setup, implementation, evaluation, and interpretation of results, were carried out by the authors. 

\section{Latin-Script Transliterations of the Case Studies}
\label{app:case-transliterations}

This section provides Latin-script transliterations of the Hindi
passages in the two case studies.
\subsection{Case I: Translation Hallucination}

\paragraph{Hindi translation.}
\texthindi{एक} 15 \texthindi{वर्षीय
\softred{\textbf{आफ्रिकान द्रीपानोसोमा संक्रमणज रोग पुरुष}} जिसका}
BMI 22 \texthindi{है, उसकी माता उसे उसके पथ्यसम्बन्धी आदतों में
परिवर्तन के बारे में चिंता व्यक्त करने के लिए उसके काया चिकित्सक
के पास लाती हैं।}

\paragraph{Latin-script transliteration.}
Ek 15-varṣīya
\softred{\textbf{āfrikān drīpānosomā saṅkramaṇaj rog puruṣ}}
jiskā BMI 22 hai, uskī mātā use uske pathya-sambandhī ādatoṃ
meṃ parivartan ke bāre meṃ cintā vyakt karne ke liye uske
kāyā cikitsak ke pās lātī haiṃ.

\subsection{Case II: Reasoning Styles}

\paragraph{Question in Hindi.}
\texthindi{एक रोगी में तेज़ बुखार, गर्दन में अकड़न और प्रकाश से डर
पाया गया है। सबसे संभावित निदान क्या है?}

\paragraph{Latin-script transliteration.}
Ek rogī meṃ tez bukhār, gardan meṃ akaṛan aur prakāś se ḍar
pāyā gayā hai. Sabse sambhāvit nidān kyā hai?

\paragraph{Answer options.}
\begin{enumerate}
    \item[(A)] \texthindi{मस्तिष्क ट्यूमर} 
    (\textit{mastiṣk ṭyūmar}; Brain tumor)
    
    \item[(B)] \texthindi{मेनिन्जाइटिस} 
    (\textit{meninjāiṭis}; Meningitis)
    
    \item[(C)] \texthindi{मिर्गी} 
    (\textit{mirgī}; Epilepsy)
    
    \item[(D)] \texthindi{माइग्रेन} 
    (\textit{māigren}; Migraine)
\end{enumerate}

\paragraph{Hindi reasoning.}
\texthindi{तेज़ बुखार और गर्दन में अकड़न मस्तिष्कावरण की सूजन का
संकेत देते हैं, जबकि प्रकाश से डर इस सूजन के कारण तंत्रिकीय
संवेदनशीलता बढ़ने से होता है। इन लक्षणों का संयोजन विकल्प}
(B) \texthindi{मेनिन्जाइटिस से मेल खाता है।}

\paragraph{Latin-script transliteration.}
Tez bukhār aur gardan meṃ akaṛan mastiṣkāvaraṇ kī sūjan kā
saṅket dete haiṃ, jabki prakāś se ḍar is sūjan ke kāraṇ
tantrikīya saṃvedanaśīlatā baṛhne se hotā hai. In lakṣaṇoṃ
kā saṃyojan vikalp (B) meninjāiṭis se mel khātā hai.

\end{document}